\documentclass[10pt]{article}
\usepackage[preprint]{tmlr}

\usepackage{amsmath,amsfonts,bm}

\def\eqref#1{equation~\ref{#1}}

\def\1{\bm{1}}

\DeclareMathAlphabet{\mathsfit}{\encodingdefault}{\sfdefault}{m}{sl}
\SetMathAlphabet{\mathsfit}{bold}{\encodingdefault}{\sfdefault}{bx}{n}

\usepackage{hyperref}
\usepackage{url}
\usepackage{graphicx}
\usepackage{amsmath,amssymb} 
\usepackage{xcolor}
\usepackage{booktabs}

\usepackage{flafter}
\newcommand{\chg}[1]{#1}
\newenvironment{ack}{}{}   
\newcommand{\rev}[1]{#1}

\title{BASILISC: Bayesian Alignment of Siamese-Integrated \\Latents for Inferring Single-cell \chg\rev{Correspondence}}

\author{\name Sepideh Saran \email sepideh.saran@mdc-berlin.de \\
      \addr The Berlin Institute for Medical Systems Biology,\\
      Max Delbr\"uck Center for Molecular Medicine, Germany\\
      Department of Electrical Engineering and Computer Science,\\
      Technical University of Berlin, Germany
      \AND
      \name Mahsa Ghanbari \email mahsa.ghanbari@mdc-berlin.de \\
      \addr The Berlin Institute for Medical Systems Biology,\\
      Max Delbr\"uck Center for Molecular Medicine, Germany\\
      Department of Biology, Humboldt University of Berlin, Germany
      \AND
      \name Uwe Ohler \email uwe.ohler@mdc-berlin.de \\
      \addr The Berlin Institute for Medical Systems Biology,\\
      Max Delbr\"uck Center for Molecular Medicine, Germany\\
      Department of Biology, Humboldt University of Berlin, Germany\\
      Department of Computer Science, Humboldt University of Berlin, Germany}

\begin{document}

\maketitle

\begin{abstract}
Single-cell assays are the established method to study the genome one cell at a time. Unimodal assays measure one molecular layer, and the newer multimodal assays measure several. Most data are single-modality, so integrating them relies on models trained on the multimodal assays, where the cell correspondence across modalities is known. These models return a matching, but rarely a confidence measure. We present BASILISC (Bayesian Alignment of Siamese-Integrated Latents for Inferring Single-cell \chg\rev{Correspondence), a framework for matching single-cell representations with uncertainty on the correspondence itself. A BASILISC model is a probabilistic siamese network over the two modalities with a predictive distribution over the correspondence for a cross-modal pair of cells. We instantiate it at two levels: at the cell-type level the question is whether the two cells share a cell type, and at the cell level which cell is the partner in the other modality.} To our knowledge, this is the first uncertainty score on a learned cross-modal correspondence evaluated against a known ground-truth pairing. BASILISC learns a biologically meaningful similarity across modalities that reflects cell lineage: an unseen cell type is matched to its lineage relative, matching still works on such held-out cell types \chg\rev{(balanced accuracy 0.86 on unseen against 0.93 on seen types, cross-batch pairs)}, and the predictive uncertainty rises on these out-of-distribution cells. The uncertainty score also selects high-quality matches for downstream multimodal analysis: discarding the least certain fifth of the pairs raises accuracy from 0.93 to 0.98. At the cell level, \chg\rev{a deep ensemble of the deterministic matcher reaches 0.062 matching accuracy on the benchmark test set, above the competition winner; on the Bayesian matcher's assignment, the posterior flags the wrong matches that cross a cell-type boundary better than a deterministic read of the same scores (AUROC 0.68 against 0.64).}
\end{abstract}
\clearpage
\section{Introduction}
\label{sec:intro}
In complex systems, a single measurement modality captures only part of an entity; to solve prediction problems that depend on the whole, the representations learned from each modality have to be brought into correspondence. The cell is such an entity, and to understand the biological processes that underlie cellular identity and disease, we need a meaningful multimodal representation of the cell state. Cross-modal matching models usually learn two encoders, one representing each data modality, and predict a point-estimate matching score for the correspondence of a given pair. Such a deterministic model can tell us which pairs are the best matches, but not how close the runner-up was, or whether the assignment would survive a different draw of the training data, or whether a data point lies where the encoders were never trained. In this work, we focus on the problem of multimodal representation matching. We place a posterior on the encoders and the uncertainty score on the correspondence itself, read the uncertainty of the assignment from the disagreement of the posterior draws, and evaluate that uncertainty where the true pairing is known: paired single-cell multi-omics data.

Single-cell assays are important means of studying the individual cell, in contrast to bulk assays, which average over all cells in a sample and thereby can report only general trends \citep{tanay2017scaling, luecken2019current}. Most single-cell datasets are unimodal, whereas multimodal datasets are more costly and have become available only recently \citep{stoeckius2017simultaneous, chen2019snare, ma2020chromatin}. A unimodal assay measures a single aspect of the cell, such as how much each gene is expressed (scRNA-seq) \citep{zheng2017massively} or which regions of the chromatin are accessible, that is, depleted of nucleosomes and therefore available for regulatory proteins to bind (scATAC-seq) \citep{buenrostro2013transposition, buenrostro2015single}. With unimodal assays, each assay is run on a separate but comparable cell population and the two are matched computationally afterward; no true cell-level correspondence exists, so the matching cannot be checked against a ground truth. Multimodal assays instead measure two or more modalities simultaneously in the same cell, yielding a paired dataset in which the modalities are linked through their cell of origin, providing us with the ground truth for the cross-modal matching task. Therefore, paired datasets generated by multimodal single-cell assays are used to train models that can, in turn, be used to integrate unpaired unimodal datasets \citep{luecken2021sandbox, lance2022multimodal}.

The matching of cell populations across modalities that have been assayed in different cells is termed diagonal integration \citep{argelaguet2021computational}. From a machine learning perspective the methods for diagonal integration fall into two categories. In the first, a representation is learned for each modality separately and the correspondence is established afterward as a separate step, by optimal transport couplings \citep{demetci2022scot, cao2022pamona, cao2020unsupervised}, mutual nearest neighbors \citep{haghverdi2018batch, stuart2019comprehensive}, or a mapping fitted between the two spaces \citep{wu2021babel}. In the second, a single embedding space is learned for both modalities at once, either by a shared latent-variable generative model such as a multimodal variational autoencoder \citep{liam, gayoso2021totalvi, ashuach2023multivi, gong2021cobolt, cao2022multi}, by matrix factorization with shared factors \citep{welch2019single, argelaguet2020mofa}, or by manifold alignment \citep{liu2019mmdma, cao2020unsupervised}, which matches the geometry of the two manifolds under objectives that preserve local neighborhood structure and require no cell-level correspondence. In both cases a correspondence score for a given cross-modal input pair is obtained from the learned space \citep{tu2022clue, gossi2023matchclot, liu2023minnet}, and in most existing methods it is obtained without measuring the confidence of the model in the predicted match score. On one benchmark dataset \citep{luecken2021sandbox} which we also use here, the strongest methods for the matching task are the models CLUE \citep{tu2022clue}, MatchCLOT \citep{gossi2023matchclot}, and the siamese matcher MinNet \citep{liu2023minnet}, all of which train two encoders with a contrastive objective and read the match off the similarity matrix, by argmax or by entropic optimal transport \citep{cuturi2013sinkhorn}. A soft transport coupling looks like a distribution over partners, but it is a normalized similarity: it carries no posterior and no calibration, and it says nothing about how far the model would have to be wrong for the match to change.

Genomic measurements are high-dimensional, sparse, and noisy. Uncertainty quantification provides a measure of confidence in the model's prediction that can reflect the shortcomings of the data. The uncertainty score indicates which predictions are reliable enough to carry into downstream analysis \citep{review1, huellermeier2021aleatoric}. Uncertainty quantification has been studied extensively for classification and regression in data-rich domains such as computer vision \citep{kendall2017uncertainties, filos2019systematic, nado2022uncertainty, arctique}, but less so in computational genomics \citep{saran2026uncertainty, engelmann2022uncertainty, bajwa2024characterizing, degu}. The standard methods place a distribution over the weights or over stochastic forward passes: deep ensembles \citep{lakshminarayanan2017simple}, Monte Carlo dropout \citep{gal2016dropout}, and mean-field variational inference \citep{blundell2015weight, wen2018flipout}, and the resulting score is used to abstain on unreliable predictions \citep{geifman2017selective} and to flag inputs from outside the training distribution \citep{hendrycks2017baseline, ovadia2019can}. In single-cell analysis, where uncertainty is reported at all, it is placed on the cell type label \citep{dedonno2023scpoli, engelmann2022uncertainty, lopezdecastro2025conformal} or on the latent representation of one cell \citep{gayoso2021totalvi, ashuach2023multivi}; none of these places it on the correspondence between two cells. Moreover, representation matching differs in that the prediction is a relation between two inputs, so the uncertainty is a property of the pair rather than of either element, and it may arise from either encoder or from the geometry of the space in which the two are compared. Related questions have been addressed in retrieval and metric learning through probabilistic embeddings and calibrated similarity scores \citep{oh2019hedged, chun2021probabilistic}, but uncertainty on a cross-modal correspondence has received limited attention, and is rarely evaluated against a known ground-truth pairing. 

We present BASILISC, Bayesian Alignment of Siamese-Integrated Latents for Inferring Single-cell \chg\rev{Correspondence}. \chg\rev{The} BASILISC \chg\rev{model} learns biologically meaningful relations between single-cell modalities and places the uncertainty score on the correspondence itself, and we ask whether that score identifies the matches a model should decline to make. We build it at two levels of the same problem: matching at the cell-type level and the cell level. At both levels we report predictive performance for the matching tasks, compare it to the published benchmark methods, and report what the uncertainty score adds: whether it separates wrong matches from correct ones, whether highly uncertain matching pairs point to underlying biology, and whether it flags a cell type the model never saw during training \chg\rev{(i.e. out-of-distribution detection)}. 

Our main contributions are threefold: First, we place an uncertainty score on the correspondence between the learned representations of two modalities, rather than on a label, a latent, or a translated profile, and evaluate it against a known ground-truth pairing; to our knowledge this has not been done before. Second, we show what \chg\rev{the uncertainty score can be useful for} at two levels of granularity, the cell type and the cell, on three data types: The score is higher on wrongly matched pairs and on cells of a type the model never saw, so a matcher can \chg\rev{filter out a low-confidence} match and flag an out-of-distribution input; every comparison is against \chg\rev{a deterministic model with equivalent architecture and} the same assignment, and we state where the score is limited. Third, we show \chg\rev{where} the \chg\rev{score} reflects a source \chg\rev{of uncertainty}: the pairs the model is unsure of are cells of one lineage family, the learned similarity follows lineage, and the correct matches with the highest uncertainty are candidates \chg\rev{to verify if the cells are} in transition, where chromatin opens ahead of transcription.

The rest of the paper is organized as follows. Section~\ref{sec:task} defines the matching task, Section~\ref{sec:method} the two BASILISC models, and Section~\ref{sec:experiments} the experimental setup. Section ~\ref{sec:results} reports our results on evaluating matching predictive performance and the uncertainty quantification at the cell-type level and at the cell level, on the multiome benchmark, with the \chg\rev{cell-type-level} replication on CITE-seq in the supplement \chg\rev{(Section~\ref{supp:cite})}. Section~\ref{sec:discussion} concludes the paper, states what the uncertainty scores on correspondences do and do not deliver, the limitations of the study, and an outlook to future research.

\section{Multimodal single-cell data integration}
\label{sec:task}

We treat diagonal integration as a supervised \chg\rev{representation matching} problem. Given a cell observed in one modality and a cell observed in another, the model predicts a score between $0$ and $1$ to indicate whether the two are the same cell, or, at a coarser level, cells of the same type. Training uses paired data, where every cell is measured in both modalities and the pairing is known; at test time the model is asked about cells it has not seen. The match score has to carry across the technical variation of a new dataset, generalize to rare and unseen cell types. \chg\rev{This matching model has two major uses} in practice: transferring labels from a reference atlas to a new dataset \chg\rev{(cell-type transfer)}, and finding, for a query cell, its counterpart in the other modality \chg\rev{(unpaired-data integration)}.

The data we use in our experiments are the multimodal single-cell benchmark of the NeurIPS 2021 integration competition \citep{luecken2021sandbox, lance2022multimodal}, a curated collection of human bone marrow mononuclear cells from ten donors profiled at four sites with two assay types that together cover three types of single cell modalities. The multiome assay (10x Genomics Chromium Single Cell Multiome) measures gene expression (GEX) and chromatin accessibility (ATAC) in the same nucleus, and the CITE-seq assay \citep{stoeckius2017simultaneous} measures gene expression and surface protein abundance (ADT) in the same cell. The three modalities carry different kinds of signals (supplement, Section~\ref{supp:data}), so an integration method needs to represent each modality on its own terms and learn a mapping between the modality-specific representations\chg\rev{, without flattening out the biological differences across modalities}. Moreover, each site–donor combination is a technical batch, and the systematic difference between batches, from sample handling, sequencing depth, and the donor, is the batch effect that every integration method has to survive: remove the batch effect while preserving biologically meaningful signal in data. A summary of both datasets is given in Table~\ref{tab:datasets} in the supplement, together with the number of cells per type and per batch (Tables~\ref{tab:cells_per_type}, \ref{tab:multiome_by_batch}, and \ref{tab:cite_by_batch}).

The \chg\rev{NeurIPS} competition's match-modality task asks, for each cell in one modality of a held-out test set, which cell in the other modality is its partner. The test set is the $20{,}009$ nuclei of one site, in three batches, and the score is the average probability the method places on the true partner (the cell level in our experiments); a one-to-one assignment scores exactly its accuracy. The winning entry, CLUE \citep{tu2022clue}, reached $0.058$, and MatchCLOT \citep{gossi2023matchclot}, published after the competition, $0.0755$; both are deterministic. We use the same test set and score at the cell level, so that our \chg\rev{ results are comparable}.

\section{Method}
\label{sec:method}
\begin{figure}[t]
  \centering
  \includegraphics[width=0.92\textwidth]{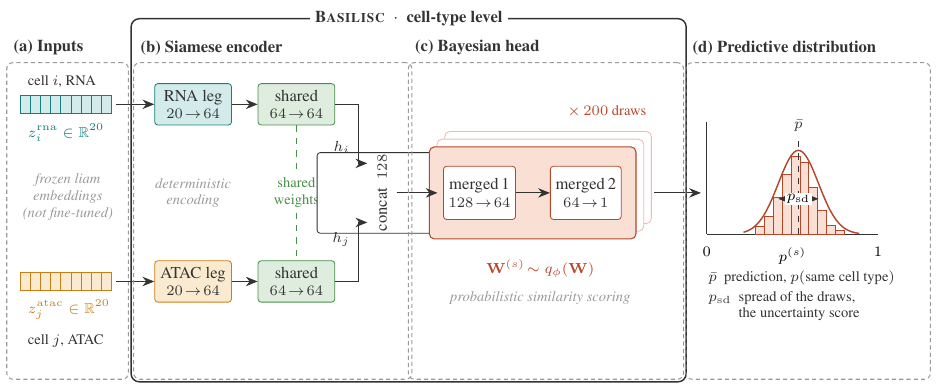}
  \caption{\textbf{\textsc{Basilisc} at the cell-type level: from frozen embeddings to a predictive distribution.} The boxed region, panels (b) and (c), is the siamese model with deterministic legs and shared layers and a probabilistic head;
(a) is the frozen embedding as input and (d) is a readout of the predictive distribution over $200$ posterior draws.}
\label{fig:overview}
\end{figure}
\begin{figure}[t]
  \centering
  \includegraphics[width=0.92\textwidth]{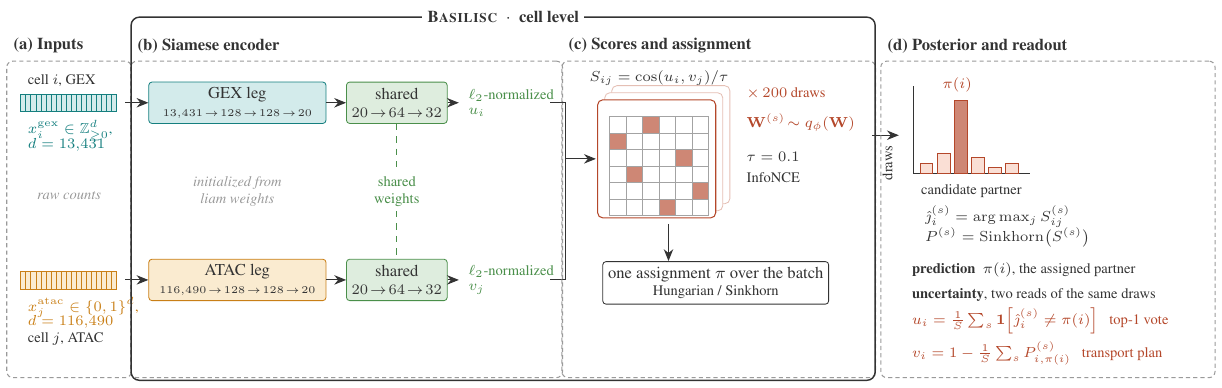}
  \caption{\textbf{\textsc{Basilisc} at the cell level: from raw counts to an assignment and its uncertainty.} The boxed region, panels (b) and (c), is the trained model, with legs initialized from single-modality Liam weights and a shared projection; (a) is the raw-count input, and (d) is the readout: one assignment $\pi$ over the batch, and the two uncertainty scores $u_i$ and $v_i$ read from 200 posterior draws per seed, pooled over the five seeds.}
\label{fig:matcher}
\end{figure}
BASILISC is a \chg\rev{framework for the diagonal single-cell data integration task at two levels. A BASILISC model is a probabilistic siamese neural} network, trained on paired single-cell data to score the correspondence between a cell observed in one modality and a cell observed in the other. We build two versions of BASILISC: one to predict the matching at cell-type level and one for cell level. We compare different methods for making BASILISC uncertainty-aware: the posterior is approximated by mean-field variational inference (MFVI, the default), by Monte Carlo dropout, or by a deep ensemble of five networks; the architecture is the same in each case, and the deterministic network is the control. The cell-type level BASILISC is a classifier over frozen single-modality embeddings that answers whether two cells are of the same type, and the cell level BASILISC matcher trained end-to-end from raw counts that answers which cell in the other modality is the same cell. Both produce, for every prediction, an uncertainty score read from the disagreement of the posterior draws, and both are compared with a deterministic version of the same network. Figures~\ref{fig:overview} and~\ref{fig:matcher} show the two architectures for the multiome data; the same architectures are used for CITE-seq, with the ATAC leg replaced by an ADT leg.

\subsection{Cell-type level: a Bayesian siamese classifier over frozen embeddings}
\label{sec:method_type}

\textbf{Inputs.} Each cell enters as a 20-dimensional embedding computed by Liam \citep{liam}, a variational autoencoder for single-cell data, trained here on one modality alone with no cell-type labels and no access to the pairing; one model for RNA and one for ATAC (RNA and ADT for CITE-seq). The embeddings are frozen and never fine-tuned, and the two latent spaces are unaligned (supplement\chg\rev{, Section~\ref{supp:embeddings}}), so the correspondence between them is entirely learned by \chg\rev{the BASILISC model}.

\textbf{Architecture} (Figure~\ref{fig:overview}). The network has one leg per modality, a shared block, and a merged head. Each leg is a \chg\rev{fully-connected} layer $20 \to 64$ with GELU activation\citep{hendrycks2016gelu} and layer normalization\citep{ba2016layer}, with its own weights, since the two inputs live in different spaces and the three modalities have very different data characteristics. The shared block, a \chg\rev{fully-connected} layer $64 \to 64$ with GELU and layer normalization, is applied to the output of both legs with the same weights, which is what makes the network siamese. The merged head concatenates the two outputs and maps them through a \chg\rev{fully-connected} layer $128 \to 64$ with GELU and a another $64 \to 1$ to a logit, whose sigmoid is the predicted probability $p$ that the two cells are of the same type.

\textbf{Posterior.} In the Bayesian version, the weights and biases of the two merged layers are replaced by a mean-field Gaussian posterior with a $\mathcal{N}(0, 1)$ prior, trained by variational inference \citep{blundell2015weight} with the local reparameterization\citep{kingma2015variational} trick. We call this the \emph{base} placement. In a second placement, \emph{legs}, every \chg\rev{fully-connected} layer of the network is variational, legs and shared block included. We also train the same architecture with Monte Carlo dropout (dropout 0.2 on the input and the hidden layer of the merged head, kept active at test time) \citep{gal2016dropout} and as a deep ensemble of five deterministic networks \citep{lakshminarayanan2017simple}, so that the posterior is compared with the  deterministic network as well as two other common approximation methods.

\chg\rev{\textbf{Prediction and uncertainty score.} At test time each pair is scored under $S = 200$ posterior draws, which gives a distribution of probabilities per pair. Its mean $\bar p$ is the prediction, called positive at $\bar p \geq 0.5$; its standard deviation $p_{\mathrm{sd}}$ is the uncertainty score. The deterministic control for every reading of $p_{\mathrm{sd}}$ is the confidence margin, $0.5 - |\bar p - 0.5|$, which is high when the prediction is close to the decision boundary and is available to the deterministic network as well; whatever $p_{\mathrm{sd}}$ does that the margin does not is what the posterior adds.}

\subsection{Cell level: a Bayesian contrastive matcher from raw counts}
\label{sec:method_cell}

\textbf{Inputs and architecture} (Figure~\ref{fig:matcher}). Each cell enters as its raw count vector\chg\rev{ from the multiome data}, 13,431 genes for GEX and 116,490 peaks for ATAC. The two legs are encoders shaped as Liam's single-modality encoders, $n_{\mathrm{in}} \to 128 \to 128 \to 20$ with layer normalization and ReLU, and they are initialized from the weights of the trained single-modality Liam models, so that training starts from an encoder that already represents each modality on its own. Above the legs sits one shared projection, $20 \to 64 \to 32$, through which both modalities pass with the same weights, as the shared block of the cell-type-level model does. Its 32-dimensional output is $\ell_2$-normalized.

\textbf{Posterior.} In the Bayesian version, every \chg\rev{fully-connected} layer of the encoders and of the shared projection is variational, with the same mean-field Gaussian, prior and training as above, except the first layer of each encoder, which stays a point estimate (supplement\chg\rev{, Section~\ref{supp:embeddings}}). \chg\rev{At inference, one posterior draw samples every variational weight once from its Gaussian and holds the sampled values fixed while the whole evaluation set is scored, so that a draw is one network, a single function applied to every cell, rather than a fresh sample per cell or per minibatch. Repeating this gives 200 draws per seed (supplement, Section~\ref{supp:scores}), and the draws can be compared as assignments.} Liam's dropout of $0.1$ is kept in both arms as a training regularizer and is off at evaluation.

\textbf{Cell partner assignment.} \chg\rev{Every experiment is repeated with five seeds, each a separately trained model that differs in the Liam encoder initialization and the training order (supplement, Section~\ref{supp:constructions}). For a test batch with $n$ GEX cells and $n$ ATAC cells, each trained model gives an $n \times n$ score matrix $S^{(k)}$, the scaled cosine similarity of every GEX cell with every ATAC cell of the batch. We average the five matrices and solve one linear assignment problem on the mean matrix $\bar S$: among all one-to-one pairings of GEX cells to ATAC cells, find the pairing $\pi$ that maximizes $\sum_i \bar S_{i,\pi(i)}$. We solve it exactly with the Hungarian algorithm \citep{kuhn1955hungarian}.} The prediction for cell $i$ is its assigned partner $\pi(i)$, and the matching accuracy is the share of cells whose assigned partner is their own\chg\rev{.}\chg\rev{ Under} the benchmark's match-modality score\chg\rev{ a} one-to-one assignment scores exactly its accuracy, so our number is read beside the published ones.

\chg\rev{\textbf{Uncertainty scores.} The posterior arm draws 200 weight samples per seed, each giving its own score matrix $S^{(s)}$; the draws of the five seeds are pooled and read on the assignment $\pi$. Two reads of the same draws give two uncertainty scores per cell. The \emph{top-1 vote} $u_i$ is the share of draws that would have matched $i$ to a cell other than $\pi(i)$. The \emph{transport-plan read} $v_i$ is one minus the mass the draws' entropic transport plans \citep{cuturi2013sinkhorn} place on $\pi(i)$; it reads the whole row of each draw, not only its argmax. Both are high when the posterior does not agree with the assignment. The deterministic control is the same transport read on the mean score matrix, $1 - P_{i,\pi(i)}$ with $P = \mathrm{Sinkhorn}(\bar S / T)$, the Sinkhorn weight: the same read of the same assignment with no posterior. Its temperature $T$ is chosen on the AUROC for any wrong assignment and never on the error type under test.} \chg\rev{Training is described in the supplement, Section~\ref{supp:training}; the formulas of the scores and the controls in Section~\ref{supp:scores}.}
\section{Experimental framework}
\label{sec:experiments}

The experiments ask\chg\rev{} three things of the uncertainty score: whether it separates wrong predictions from correct ones, whether \chg\rev{filtering} on it removes the errors that matter, and whether it is high on cells of a type the model never trained on. Each question is asked under a controlled construction of the data, and each answer is read beside the deterministic control, so that what the posterior adds is separated from what the network already knew.

\textbf{Arms.} \chg\rev{At the cell-type level, five arms share one architecture (Section~\ref{sec:method_type}) and differ only in the posterior:}
\begin{enumerate}
\item \chg\rev{\textbf{BNN}: a mean-field Gaussian posterior, fitted by MFVI, on the two layers of the merged head only; legs and shared block stay point estimates (the \emph{base} placement of Section~\ref{sec:method_type}); the default.}
\item \chg\rev{\textbf{BNN, every layer}: the same posterior on every fully-connected layer, legs and shared block included (the \emph{legs} placement).}
\item \chg\rev{\textbf{MCD}: Monte Carlo dropout, rate 0.2 on the input and the hidden layer of the merged head, kept active at test time.}
\item \chg\rev{\textbf{ENS}: a deep ensemble of five deterministic networks trained from different initializations.}
\item \chg\rev{\textbf{DET}: the deterministic network, the control; its uncertainty score is the confidence margin.}
\end{enumerate}
\chg\rev{Three baselines use the same frozen embeddings and the same pair labels with no network: (6) $k$-nearest-neighbor label transfer, (7) an orthogonal Procrustes rotation with Platt scaling, and (8) a logistic pair classifier (supplement, Section~\ref{supp:scores}). At the cell level there are two arms, (1) the MFVI matcher (BNN) and (2) the deterministic matcher (DET), each read as a single seed and as a five-seed ensemble, and compared with the published scores of CLUE \citep{tu2022clue} and MatchCLOT \citep{gossi2023matchclot} on the same test set.}

\textbf{Data constructions.} \chg\rev{We train and evaluate on constructions of the same data that differ in what is withheld at training; each asks one question of the uncertainty. At the cell-type level, train and test are built from cells, not from the given pairing: for each cell type we draw up to 100 cells, split them 70/15/15 into train, validation and test, and form every positive pair (two cells of the type) and as many negative pairs (a cell of the type with a cell of another type) within each split, so that train, validation and test share no cells and every type is balanced. Pairs are built in two populations. \emph{Cross-batch} pairs ignore the batch, so the two cells may come from different sites and donors; \emph{within-batch} pairs take both cells from the same batch. The reason is the batch effect: cells measured together resemble each other in the embedding whatever their type, so a model could score a pair by whether the two cells share a batch instead of a cell type. Evaluating on within-batch pairs makes sure we are not picking up the batch effect, and a model that does not use the batch scores the two populations alike. We report cross-batch pairs unless stated otherwise. At the cell level, the model is trained and tested on the NeurIPS 2021 competition split, with the competition's own test set, so that our matching accuracy is read beside the published scores of the competition's methods; this split exists for the multiome data only. At the cell-type level:}
\begin{enumerate}
\item \chg\rev{\textbf{mix}: every cell type and batch is seen in training; the test pairs are new cells of seen types. The in-distribution reference: does the uncertainty find the model's own errors?}
\item \chg\rev{\textbf{LOCTO} (leave-one-cell-type-out): one cell type is removed from training and appears only at test time, while its close relatives stay in training. Does the uncertainty rise on a type the model never saw when a near neighbor it knows is available? This is the hardest novelty case.}
\item \chg\rev{\textbf{LOFO} (leave-one-family-out): a whole lineage family (Table~\ref{tab:families}) is removed, so the held-out types have no close relative in training. Does the model abstain, rather than transfer, when nothing related was seen?}
\item \chg\rev{\textbf{LTFO} (leave-two-families-out): two families are removed, so the test set contains pairs the model has seen from neither side. Does the model still tell the two unseen families apart, or are all unseen cells one blob to it?}
\item \chg\rev{\textbf{few-shot}: LOFO with $k \in \{0, 5, 10, 20, 30, 50\}$ cells of each type of the held-out family returned to training. How many cells, and which, bring a family back?}
\end{enumerate}
\chg\rev{At the cell level:}
\begin{enumerate}
\item \chg\rev{\textbf{challenge}: the NeurIPS 2021 competition split; the 20{,}009 site-4 nuclei are the test set, every other cell is available for training and validation. Accuracy beside the published methods, and which errors the uncertainty sees.}
\item \chg\rev{\textbf{challenge-LOCTO}: the same split with one of eight cell types removed from training; its held-out cells are matched against the seen cells of the same batch. The LOCTO question at the level of the cell.}
\end{enumerate}
\chg\rev{Every construction is run for five seeds. The full definitions, the held-out lists and the pair populations are given in the supplement, Section~\ref{supp:constructions}.}

\section{Results}
\label{sec:results}

\chg\rev{At the cell-type level, every number is a mean over five seeds on the multiome data unless stated; the held-out designs pool 22 cell types $\times$ five seeds $\times$ two pair populations, 220 runs. At the cell level, every number is on the 20{,}009 test cells of the NeurIPS 2021 competition split, five seeds, 200 posterior draws per seed. The AUROC of an uncertainty score is the AUROC of that score used as the ranking variable for a binary label, with the uncertainty as the positive direction. We report it for two labels: wrong versus correct prediction (error detection) and held-out versus seen cell (novelty detection, on positive pairs at the cell-type level). Each score is read beside its deterministic control, the confidence margin at the cell-type level and the Sinkhorn transport weight at the cell level (supplement, Section~\ref{supp:metrics}), so that what the posterior adds is separated from what the network already knew. For every analysis we wrote down the expected outcome and the bar before running it, in a dated log released with the code.}

\subsection{Cell-type level}
\label{sec:results_type}

\paragraph{\chg\rev{Classification.}}
\chg\rev{On pairs of seen cell types (mix) the model reaches AUROC 0.976 $\pm$ 0.002 on cross-batch pairs and 0.977 $\pm$ 0.008 within batch, with accuracy 0.932, from about 70 training cells per type (Table~\ref{tab:methods}). Cross-batch and within-batch pairs are classified alike, so the model is not using the batch. Holding one type out leaves the seen types at 0.975 $\pm$ 0.006. Most of this accuracy is in the frozen embeddings: the linear pair classifier on the same inputs reaches accuracy 0.917, and on pairs that contain the held-out type it matches the network (balanced accuracy 0.882 against 0.861). On seen types, what the network adds is a better probability: its Brier score and NLL are below those of every baseline, in 5 of 5 seeds (Table~\ref{tab:methods}). On unseen types, what it adds is the uncertainty, which the rest of this section is about.}

\paragraph{\chg\rev{Error detection and filtering.}}
\chg\rev{On the mix runs we rank the test pairs by $p_{\mathrm{sd}}$ and ask whether the wrong predictions sit at the top. They do. Per run, $p_{\mathrm{sd}}$ separates misclassified pairs from correct ones at AUROC 0.856 $\pm$ 0.016; median $p_{\mathrm{sd}}$ is 0.010 on correct pairs and 0.079 on errors. Figure~\ref{fig:risk_coverage}A shows what filtering on it does: as the least certain pairs are discarded, accuracy on the kept pairs rises in every run. Discarding the least certain fifth lowers the error rate from 6.8\% to 2.5\%, about 70\% of the errors, and the gain from keeping the most certain half is positive in all ten runs (Figure~\ref{fig:risk_coverage}B). The confidence margin of the deterministic network, which needs no posterior, does the same: 0.858 against 0.856, with the same curve to the third decimal (dashed line in Figure~\ref{fig:risk_coverage}A). This is expected for a binary classifier. With $\sigma_z$ the standard deviation of the logit over the draws, $p_{\mathrm{sd}} = \bar p(1-\bar p)\,\sigma_z$ to first order, so the predictive SD is largest where the prediction is closest to 0.5, which is exactly what the margin measures; per pair the two correlate at Spearman 0.95 to 0.99 at every posterior width we trained (Table~\ref{tab:kl}). On seen types, the posterior finds the errors, and so does the margin.}

\begin{figure}[t]
\centering
\includegraphics[width=\textwidth]{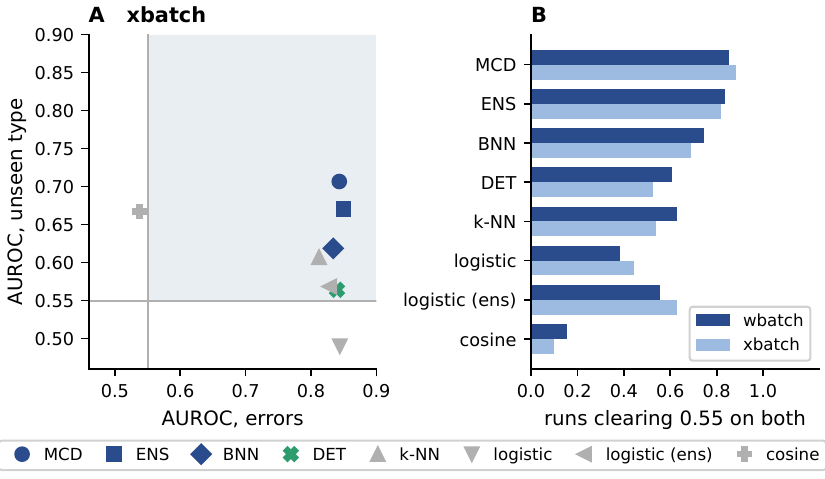}\\[2pt]
\includegraphics[width=\textwidth]{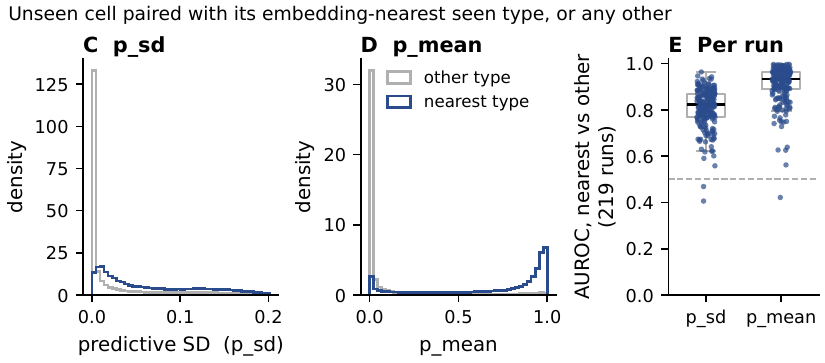}
\caption{Uncertainty at the cell-type level. (A) Each method's own uncertainty score on the 110 cross-batch LOCTO runs: AUROC for detecting its own errors on seen-type pairs against AUROC for detecting the held-out type's positives among seen positives. The shaded region is above 0.55 on both. (B) The share of runs in which one score clears 0.55 on both jobs, for both pair populations. (C--E) An unseen type's negatives, split by whether the partner is the seen type nearest to it in the input embedding or any other seen type: (C) predictive SD and (D) predictive mean of the two groups, pooled over 219 runs (one run has no nearest-type pair); (E) per-run AUROC of each score for separating the two groups, one point per run, dashed line at chance.}
\label{fig:type_level}
\end{figure}

\paragraph{\chg\rev{A cell type never seen in training.}}
\chg\rev{In LOCTO the model is trained without one cell type and then scores pairs that contain it. It still classifies them. On the held-out type's own pairs (47\% positive) AUROC is 0.920 $\pm$ 0.057 against 0.975 on the seen types of the same runs, and recall 0.816 against 0.947: a cell of an unseen type is matched to its own type most of the time, because a relative of it was in training. The question is whether the model knows that it has not seen the type. It does, through $p_{\mathrm{sd}}$: per run, $p_{\mathrm{sd}}$ ranks the unseen type's positives above seen positives at AUROC 0.627 $\pm$ 0.138, above 0.5 in 19 of 22 cell types. The deterministic network's margin gives 0.564 and 0.582 on the two pair populations, and every arm with a posterior (MFVI, MC dropout, ensemble) beats it ($t$ from 3.8 to 11.5). Figure~\ref{fig:type_level}C--E shows where this uncertainty sits. We took every negative pair of an unseen type and split it by the partner: the seen type nearest to the unseen one in the input embedding, or any other seen type. For the nearest type the predictive SD is spread out and the predictive mean sits away from zero (panels C and D); for every other type both collapse to zero: median $p_{\mathrm{sd}}$ is 0.057 against the nearest type and 0.0015 against any other, a 38-fold ratio. Per run, $p_{\mathrm{sd}}$ separates the two groups at AUROC 0.81 $\pm$ 0.08 (panel E). The model is certain about what the unseen cell is not. It is unsure only about the cell's relative. For label transfer from an atlas this is the case that matters: a cell of a type the atlas lacks gets its nearest relative's label with a raised uncertainty, so the transfer can be flagged for review instead of accepted or refused.}

\paragraph{\chg\rev{One score for both jobs.}}
\chg\rev{On seen types the margin ties the posterior; on an unseen type the posterior wins. Figure~\ref{fig:type_level}A puts the two jobs on one plot, one point per method, each with its own uncertainty score, over the 110 LOCTO runs per pair population (Table~\ref{tab:bothjobs}). On error detection every network-based method sits at 0.83 to 0.87; on novelty detection they spread out: MC dropout 0.707, the five-member ensemble 0.671, MFVI 0.619, and the deterministic network 0.564. The cosine similarity of the frozen embeddings flags unseen types at 0.667, so we do not claim that the posterior flags novelty better than every alternative; it is at chance on its own errors (0.54), and the linear pair classifier is the mirror image, good on its errors and at chance on novelty (0.49). Each baseline does one job. Panel B counts, per method, the runs in which the same score reaches an AUROC above 0.55 on both tasks. The three posteriors lead, 88\% and 85\% for MC dropout, 82\% and 84\% for the ensemble, 69\% and 75\% for MFVI; the deterministic network reaches 53\% and 61\% and the best baseline, $k$-NN label transfer, 54\% and 63\%. The deterministic arm is the same network with sampling switched off, so the gap between 53\% and 69\% is the posterior alone.}

\paragraph{\chg\rev{Where the posterior sits.}}
\chg\rev{Two experiments in the supplement say what the posterior's novelty signal depends on. We scaled the KL term of the loss by a factor $\lambda$, from 0, where the posterior collapses onto its mean, to 1, the standard ELBO, and beyond, which moves the posterior width over a 200-fold range. Classification does not move; novelty detection rises from 0.584 with the collapsed posterior to 0.627 at $\lambda = 1$ and saturates above half that width (Section~\ref{supp:kl}, Table~\ref{tab:kl}). Placing the same posterior on every layer instead of the merged head alone leaves classification and error detection unchanged and raises novelty detection from 0.627 to 0.683, $+0.056$ [$+0.036$, $+0.076$] over 220 paired runs. The width of the posterior and the layers it covers change what it carries about unseen types, and nothing else.}

\paragraph{\chg\rev{Lineage.}}
\chg\rev{The LOCTO runs also give a $22 \times 22$ matrix: entry $(T, X)$ is the mean predicted probability that a cell of the held-out type $T$ and a cell of the seen type $X$ share a type (Figure~\ref{fig:similarity}). It reads as a lineage tree. The T cells form one block, the B cells another, the monocytes, the progenitors and the erythroid chain their own; every held-out type's nearest neighbor among the seen types (the marked square in each row) is a lineage relative. A type the model never saw is matched to its family. The model keeps the structure of the embedding rather than sharpening it: against the cosine similarity between type centroids in the input embedding, Spearman is $+0.68$ over the 462 type pairs and the nearest neighbor is the same for 16 of 22 types; when each row of the embedding's matrix is sharpened to the entropy of the model's row, so that the two are compared at equal peakiness, the model and the embedding put the same share of a row on the held-out type's own family (0.573 against 0.568, $p = 0.48$). The uncertainty follows the same tree. An unseen cell is uncertain only against its relative (Figure~\ref{fig:type_level}C--E), and a wrong match at this level is a match to a relative. Cells with the same label are not all in one state. Chromatin opens ahead of transcription along a differentiation path \citep{ma2020chromatin, martin2021chromatin}, so two cells of one type can differ in how far accessibility has run ahead of expression. The correct matches the model is least sure of are those of the types it partly confuses with a neighboring type; one reading, which we offer as an interpretation, is that the uncertainty picks up cells in transition.}

\paragraph{\chg\rev{A held-out family, and how many cells bring it back.}}
\chg\rev{LOFO removes the relatives too. Without them the model abstains: recall on the family's positives falls from 0.82 to 0.49 and the median $\bar p$ from 0.86 to 0.48, so the held-out family's pairs land at the decision threshold, and the novelty AUROC of $p_{\mathrm{sd}}$ rises from 0.63 to 0.73. Few-shot then asks how many cells undo this. Figure~\ref{fig:fewshot}A shows recall on the family's positives against the number $k$ of its cells returned to training: it climbs steeply and reaches 0.93 at $k = 20$ on both pair populations, against 0.94 on seen types in the same runs (dashed line), and adding more cells stops helping between 20 and 30. Which cells are returned does not matter: choosing the $k$ cells closest to a seen type in the embedding is worse than a random choice by 0.015 recall, and choosing the $k$ cells farthest apart within the type is no better ($p = 0.40$). Twenty cells recover a family. Every claim of this level replicates on the CITE-seq data, 38 cell types, 390 runs, at AUROC 0.970 and 0.963 on the two pair populations (supplement, Section~\ref{supp:cite}, Table~\ref{tab:cite}).}

\subsection{Cell level}
\label{sec:results_cell}

\paragraph{\chg\rev{Accuracy, and error structure.}}
\chg\rev{On the competition's test set, the deep ensemble matches 0.0624 of the cells to their own partner, 95\% CI [0.0591, 0.0656], above CLUE's 0.058 \citep{tu2022clue} with the interval clear of it; MatchCLOT's 0.0755 \citep{gossi2023matchclot} is not reached. Five MFVI seeds averaged reach 0.0478 [0.0447, 0.0507]: at the cell level the mean-field posterior costs accuracy. An accuracy of 0.06 means that 94\% of the assignments are wrong, so the first question is where the wrong partners go. Figure~\ref{fig:errors_where} answers it for both arms; the control is the same share when every cell's partner is replaced by a random cell of the same batch. Almost nine in ten wrong assignments land on another cell of the query's own type (88.8\% and 88.4\%, against 14.3\% permuted), and of the ones that leave the type, three in four go to a lineage relative (76.2\% and 74.6\%, against 15.3\% permuted). The two arms are indistinguishable here: the error structure is the embedding's, not the posterior's. An accuracy of 0.06 at the level of the cell sits on near-perfect matching at the level of the type.}
\begin{table}[htbp]
\centering
\caption{Detecting a wrong assignment, by the kind of error. AUROC of each score for separating wrong assignments of the given kind from correct matches, Bayesian arm, pooled over the three test batches of the competition split (20,009 cells; 19,053 wrong, of which 16,852 same-type and 2,201 cross-type; of the cross-type errors 1,642 go to a lineage relative and 559 to another family).}
\label{tab:harmful}
\resizebox{\textwidth}{!}{\begin{tabular}{lccccc}
\toprule
score & any wrong & same-type & cross-type & \multicolumn{2}{c}{cross-type, partner is} \\
 & & & & a relative & another family \\
\midrule
transport-plan posterior $v_i$ & 0.804 & 0.821 & 0.679 & 0.676 & 0.686 \\
top-1 vote $u_i$ & 0.777 & 0.790 & 0.678 & 0.672 & 0.697 \\
Sinkhorn weight, best $T$ ($T=0.5$) & 0.792 & 0.812 & 0.635 & 0.646 & 0.603 \\
Sinkhorn weight, $T=1$ & 0.785 & 0.804 & 0.639 & 0.641 & 0.631 \\
seed stability & 0.701 & 0.707 & 0.656 & 0.654 & 0.662 \\
row margin & 0.604 & 0.602 & 0.618 & 0.606 & 0.652 \\
\bottomrule
\end{tabular}}
\end{table}

\paragraph{\chg\rev{Which errors the posterior sees.}}
\chg\rev{Because nearly every error stays in the type, the errors are not equally harmful. A wrong partner of the same type carries the right label into anything downstream; a wrong partner of another type carries the wrong one. We therefore read error detection by kind of error, with the bar set in advance at $+0.02$ AUROC over the control (Table~\ref{tab:harmful}, Figure~\ref{fig:errors_harmful}). Pooled over all wrong assignments, the transport-plan posterior $v_i$ reaches 0.804 against 0.792 for the Sinkhorn weight of the same mean matrix at its best temperature, $+0.013$ [$+0.009$, $+0.017$]: above zero, under the bar. Split by kind, the picture changes. On the 11\% of errors that cross types, the posterior leads in every batch, $u_i$ against the control $+0.060$, $+0.031$, $+0.027$, all above the bar, and still leads when the control's temperature is chosen on this target. On the 89\% that stay in the type it does not ($-0.007$ to $-0.043$): a wrong partner of the same type receives nearly as much weight in the transport plan as the right one, so the deterministic plan is confident either way, and this is where its lead on the pooled reading comes from. The posterior trades the harmless errors for the harmful ones. Filtering on $u_i$ and keeping the most certain half of the assignments removes 32.8\% of the cross-type errors; filtering on the Sinkhorn weight removes 25.1\%. The harmful errors are also the confident errors: ranking cross-type errors against same-type errors by uncertainty gives an AUROC below 0.5 for every score, 0.35 for the posterior. Figure~\ref{fig:errors_distance} splits the cross-type errors once more, by whether the wrong partner is a lineage relative or belongs to another family. The posterior's lead grows with the distance of the wrong partner: on errors to a relative $u_i$ reaches 0.672 against 0.646 for the control, on errors to another family 0.697 against 0.603. The control is more confident about a wrong match to another family than to a relative; the posterior is less.}

\paragraph{\chg\rev{A cell type never seen in training, at the cell level.}}
\chg\rev{Challenge-LOCTO repeats the LOCTO test with the matcher: one of eight cell types is removed from training and its test cells are matched against the seen cells of the same batch. The matching transfers: held-out cells still get a same-type partner in 0.60 to 0.96 of cases, within 0.05 of the model trained on all types in the deterministic arm and within 0.11 in the Bayesian arm. Figure~\ref{fig:locto_cell} shows the novelty AUROC per held-out type, ordered by the number of held-out test cells, for the Sinkhorn weight and the two posterior scores (50 draws per seed). For the five types with at least 751 test cells all three scores separate held-out from seen cells, at 0.67 to 0.85 for the Sinkhorn weight; for pDC, CD4$^+$ T naive and HSC the Sinkhorn weight is below chance and the posterior scores are at or below it. The size of the held-out type drives this. A small type has few same-type competitors in its batch, so its cells are assigned with more confidence than seen cells, and a confident assignment is read as a seen cell. The posterior adds $+0.016$ on average over the control, above the bar in 3 of 8 types. None of the scores we tested removes the dependence on type size, so the novelty claim of this paper is made at the cell-type level.}

\section{Discussion}
\label{sec:discussion}

We put uncertainty on the correspondence in a multimodal matching model, at two levels of granularity, the cell type and the cell, on three single-cell data types, and asked where it works and where it is limited. At the cell-type level one score is higher on the pairs the model gets wrong and on the cells of a type it never trained on, and no deterministic baseline does both\chg\rev{ as often}; an unseen type is matched to its lineage relative and is uncertain only against that relative, so the uncertainty reflects the cell lineage structure the model has learned without ever seeing it. At the cell level, the posterior flags the wrong matches that cross a type boundary, the harmful ones, better than a deterministic read of the same scores, and its lead grows with the distance of the wrong partner; \chg\rev{the deterministic ensemble reaches the published methods' accuracy, the mean-field matcher sits below it.} 
Where it is limited: on seen types the predictive SD ranks pairs as the confidence margin does, so abstaining on it works but is not a gain over a deterministic network. The separation of an unseen type is above chance in 19 of 22 types and rises when its whole lineage is absent, but it is modest, and MC dropout and a plain cosine similarity flag unseen types as well or better. At the cell level the score does not detect a never-seen type, because a small held-out type has few competitors in its batch, and on the errors that stay inside a type the deterministic transport plan is the better detector. Our empirical framework comes with limitations, which we set out here. The cell-type level works on frozen embeddings, so what the model learns about lineage is the embedding's structure carried into a correspondence, and the classification itself is within reach of a linear map on the same inputs. The Sinkhorn control's temperature is chosen on the test set, which favors the control. Our study covers three single-cell modalities, but \chg\rev{it is} limited to one tissue type, which we leave for future studies. 
\chg\rev{What we would do next. The posterior should sit on the encoders as well as on the correspondence head, because on every layer it raised novelty detection at no cost to classification. A novelty score for one-to-one assignment has to correct for the number of competitors a cell has. And the correct matches with a high uncertainty are where we would look for cells in which chromatin and transcriptome part ways.}

\subsubsection*{AI use statement}

The research question, the model design and the first implementation predate this work and were done completely without AI assistance. That earlier version was written manually in TensorFlow Probability and run on a subset of the Multiome data. From that code, we used AI assistants to port the implementation to PyTorch, refactor it, and scale it to the full datasets. The cell-level model and most of the experiments reported here were implemented the same way. We also used AI assistants for feedback on experimental design, for computing the summary statistics and tables in the supplement, for searching for related work, for producing and editing the figures. The main manuscript text is written manually and polished, and complemented by AI. Parts of the supplement text are generated by AI and verified manually. An author was in the loop at every step of any task mentioned here where AI was used, and we reviewed all AI-assisted work. 
We did not use generative AI to generate synthetic research data, scope the project, develop theoretical models, design the research question, design the paired dataset generation settings, design experimental scenarios, design Siamese model architecture. 

\subsubsection*{Ethics statement}
The aim of this work is to learn a richer multimodal representation of the cell, which can in turn help in understanding biological processes and disease mechanisms. We foresee no ethical issue with the present work. The data are the publicly released benchmark datasets described in Section~\ref{sec:task}; no personal data were processed.

\subsubsection*{Reproducibility statement}
Every reported number in the paper is produced by code that will be made available in the accompanying code base at the time of publication. Random seeds are fixed and reported for all experiments for reproducibility. Every analysis carried unit tests, and each test was verified to break on a deliberately broken input. Experiments were run on a shared Linux HPC cluster under Slurm, on NVIDIA GPUs \chg\rev{(V100, T4 and A40, 16 to 45 GB)}; one cell-type-level model trains in about one GPU-minute and one cell-level model in \chg\rev{11} to 45 minutes, depending on the card, so the full set of runs is reproducible on a single GPU within days.

\begin{ack}
\paragraph{Acknowledgments}
S. Saran and U. Ohler were supported by the Initiative and Networking Fund of the Helmholtz Association (Helmholtz AI project UNITY, ZT-I-PF-5-149) as well as by the Helmholtz-Einstein International Berlin School in Data Science (HEIBRiDS). M. Ghanbari and U. Ohler were supported by DFG research unit DeSBI (KI-FOR 5363).
\end{ack}

\bibliography{iclr2027_conference}
\bibliographystyle{iclr2027_conference}

\clearpage
\appendix
\setcounter{figure}{0}\renewcommand{\thefigure}{S\arabic{figure}}
\setcounter{table}{0}\renewcommand{\thetable}{S\arabic{table}}
\begin{center}
  {\Large\bfseries Supplementary Material}\\[4pt]
  {\large for BASILISC: Bayesian Alignment of \\Siamese-Integrated Latents for Inferring Single-cell \chg\rev{Correspondence}}
\end{center}
\vspace{1em}

\section{Data}
\label{supp:data}
Here we provide more information about the datasets used in this work. The cell-type-level experiments on CITE-seq used the competition's phase-2 file ($81{,}241$ cells), from which the single-modality embeddings were computed\chg\rev{; the tables below are from the GEO release GSE194122 ($90{,}261$ cells), which has the same annotation and batches}. Table~\ref{tab:cells_per_type} gives the number of cells per cell type in each dataset, and Tables~\ref{tab:multiome_by_batch} and~\ref{tab:cite_by_batch} the number of cells per cell type and batch.

The three modalities have different kinds of signals. Gene expression is a count of transcripts per gene, a non-negative integer that is zero for most genes in a cell and follows an overdispersed count distribution for the rest \citep{zheng2017massively, svensson2020droplet}. Chromatin accessibility is a count of sequencing fragments per peak, and because a diploid nucleus carries only two copies of each region, the count is almost always 0, 1 or 2 and the matrix is far sparser than the expression matrix; in practice it is read as binary, open or closed \citep{cusanovich2015multiplex, buenrostro2015single}. The surface protein abundance from CITE-seq is a count of antibody tags per protein over a small panel, dense rather than sparse, with a nonzero background from unbound antibodies, and is normalized by a centered log-ratio transform before use \citep{stoeckius2017simultaneous}.

\begin{figure}[htbp]
  \centering
  \includegraphics[width=\textwidth]{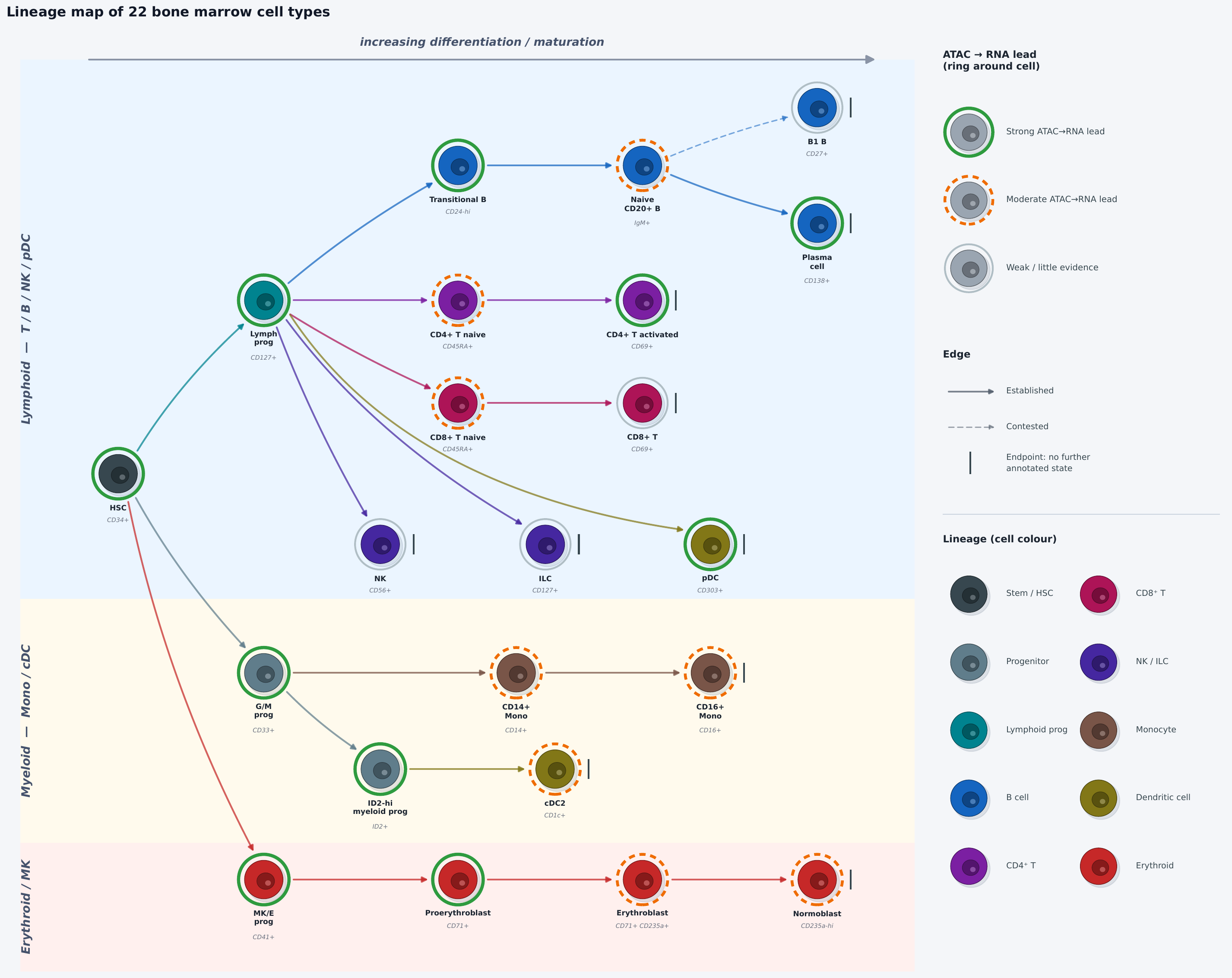}
  \caption{\textbf{Lineage map of the 22 bone marrow cell types present in the multiome data, drawn from the literature.}
  Arrows run in the direction of increasing differentiation. Solid arrows are established transitions, dashed arrows are contested, and a bar marks a type with no further annotated state in this annotation. The three background bands are the broad compartments: lymphoid, myeloid, and erythroid,/MK. The literature does not agree on every transition, and some of the placements are still debated; therefore, the map can be wrong in some details.}
\label{fig:lineage}
\end{figure}

Figure~\ref{fig:lineage} shows a lineage map for the 22 bone marrow cell types present in the multiome data. The map is drawn from the literature, not from the data, and no result depends on it. It is included as a reference for which held-out type has a close relative among the types the model still sees, which is what separates the leave-one-cell-type-out and leave-one-family-out settings. The topology follows the standard hematopoietic hierarchy \citep{ranzoni2021integrative}: the plasmacytoid dendritic cell branch and its progenitors \citep{rodrigues2018distinct, rodrigues2020novel}, the innate lymphoid and natural killer lineages \citep{ishizuka2016innate, cichocki2019human, ding2024distinct}, the monocyte subsets \citep{patel2017fate}, the erythroid chain from proerythroblast to normoblast \citep{an2014global, ludwig2019transcriptional, huang2020putative}, and the B-cell subsets, including the disputed human B1 B population \citep{martinezgutierrez2026human}. The annotations of chromatin accessibility changing ahead of transcription rest on reports of multilineage priming in hematopoietic stem cells \citep{martin2021chromatin}, of chromatin potential in paired profiles \citep{ma2020chromatin}, and of chromatin opening during T-cell activation and B-cell memory formation \citep{yukawa2020ap1, shao2024epigenetic}. These sources do not agree with each other on every transition, and some of the placements are still debated in the field; the map is our reading of them, it can be wrong in its details, and we use it only for the coarse question of which types are relatives.

Our cell-type family table (Table~\ref{tab:families}) is coarser than the map and differs from it in four placements. MK/E prog is a progenitor for us and erythroid on the map. ID2-hi myeloid prog is mono/DC for us and a progenitor on the map. Lymph prog is B for us and a root of its own on the map. pDC is a singleton for us and a dendritic cell on the map. The map also splits T into CD4$^+$ and CD8$^+$ where we keep one T family. The table was fixed before any leave-one-family-out result, and we report it as it stands, rather than redefining it to match the map. The CITE-seq annotation is finer, and its table places each of the 38 types in the same seven families, so the family-level settings hold out the same lineages on both datasets.
\begin{table}[htbp]
\centering
\caption{The two datasets of the NeurIPS 2021 multimodal single-cell benchmark used in this work. Multiome: gene expression and chromatin accessibility in the same nucleus. CITE-seq: gene expression and surface protein abundance in the same cell. A batch is one site--donor combination. The cell-level test set is the competition's test set\chg\rev{, 20,009 nuclei drawn from the three batches of site 4; it exists for the multiome data only}.}
\label{tab:datasets}
\begin{tabular}{lcc}
\toprule
 & Multiome & CITE-seq \\
\midrule
Modalities & GEX, ATAC & GEX, ADT \\
Cells & 69,249 & 90,261 \\
Features & 13,431 genes, 116,490 peaks & 13,953 genes, 134 proteins \\
Annotated cell types & 22 & 45 \\
Types used & 22 & 38 ($\geq 200$ cells) \\
Batches (site $\times$ donor) & 13 & 12 \\
Sites / donors & 4 / 10 & 4 / 9 \\
Cells per type, min--max & 108--11,589 & 18--21,693 \\
Types absent from $\geq 1$ batch & 10 of 22 & 24 of 45 \\
Cell-level test set & 20,009 \chg\rev{(from s4d1, s4d8, s4d9)} & \chg\rev{--} \\
\bottomrule
\end{tabular}
\end{table}

\begin{table}[htbp]
\centering
\caption{The lineage families used for the leave-one-family-out, leave-two-families-out and few-shot settings, and for reading where a wrong match goes. The multiome table partitions the 22 annotated types into seven families; the CITE-seq table partitions the 38 types with at least 200 cells into the same seven families.}
\vspace{6pt}
\label{tab:families}
\small
\setlength{\tabcolsep}{4pt}
\begin{tabular}{llp{0.72\textwidth}}
\toprule
\multicolumn{3}{l}{\textbf{Multiome}, 22 types} \\
Family & $n$ & Cell types \\
\midrule
T & 4 & CD4+ T naive, CD8+ T naive, CD4+ T activated, CD8+ T \\
innate lymphoid & 2 & ILC, NK \\
B & 5 & Lymph prog, Transitional B, Naive CD20+ B, B1 B, Plasma cell \\
mono/DC & 4 & CD14+ Mono, CD16+ Mono, ID2-hi myeloid prog, cDC2 \\
progenitor & 3 & HSC, G/M prog, MK/E prog \\
erythroid & 3 & Proerythroblast, Erythroblast, Normoblast \\
pDC & 1 & pDC \\
\midrule\addlinespace[8pt]
\multicolumn{3}{l}{\textbf{CITE-seq}, 38 types} \\
Family & $n$ & Cell types \\
\midrule
T & 14 & CD4+ T naive, CD4+ T activated, CD4+ T activated integrinB7+, T reg, CD8+ T naive, CD8+ T CD49f+, CD8+ T CD69+ CD45RA+, CD8+ T CD69+ CD45RO+, CD8+ T TIGIT+ CD45RA+, CD8+ T TIGIT+ CD45RO+, CD8+ T CD57+ CD45RA+, CD8+ T CD57+ CD45RO+, MAIT, gdT CD158b+ \\
innate lymphoid & 4 & ILC, ILC1, NK, NK CD158e1+ \\
B & 9 & Lymph prog, Transitional B, Naive CD20+ B IGKC+, Naive CD20+ B IGKC-, B1 B IGKC+, B1 B IGKC-, Plasmablast IGKC+, Plasma cell IGKC+, Plasma cell IGKC- \\
mono/DC & 3 & CD14+ Mono, CD16+ Mono, cDC2 \\
progenitor & 3 & HSC, G/M prog, MK/E prog \\
erythroid & 4 & Proerythroblast, Erythroblast, Normoblast, Reticulocyte \\
pDC & 1 & pDC \\
\bottomrule
\end{tabular}
\end{table}

\begin{table}[p]
\centering
\caption{Cells per annotated cell type in the two datasets, sorted by count. Multiome: 69,249 nuclei, 22 types. CITE-seq: 90,261 cells, 45 types; the seven types marked $\dagger$ have fewer than 200 cells and are excluded from the cell-type-level experiments, leaving 38.}
\label{tab:cells_per_type}
\small
\begin{tabular}[t]{lr}
\toprule
\multicolumn{2}{c}{Multiome} \\
Cell type & Cells \\
\midrule
CD8+ T & 11,589 \\
CD14+ Mono & 10,843 \\
NK & 6,929 \\
CD4+ T activated & 5,526 \\
Naive CD20+ B & 5,052 \\
Erythroblast & 4,916 \\
CD4+ T naive & 4,398 \\
Transitional B & 2,810 \\
Proerythroblast & 2,300 \\
CD16+ Mono & 1,894 \\
B1 B & 1,890 \\
Normoblast & 1,780 \\
Lymph prog & 1,779 \\
G/M prog & 1,203 \\
pDC & 1,191 \\
HSC & 1,072 \\
CD8+ T naive & 1,012 \\
MK/E prog & 884 \\
cDC2 & 859 \\
ILC & 835 \\
Plasma cell & 379 \\
ID2-hi myeloid prog & 108 \\
\midrule
Total & 69,249 \\
\bottomrule
\end{tabular}
\hspace{1em}
\begin{tabular}[t]{lr}
\toprule
\multicolumn{2}{c}{CITE-seq} \\
Cell type & Cells \\
\midrule
CD14+ Mono & 21,693 \\
CD4+ T activated & 6,966 \\
CD4+ T naive & 5,897 \\
NK & 5,434 \\
Reticulocyte & 4,272 \\
Erythroblast & 4,039 \\
Naive CD20+ B IGKC+ & 3,990 \\
CD8+ T naive & 3,107 \\
CD16+ Mono & 2,635 \\
NK CD158e1+ & 2,167 \\
Naive CD20+ B IGKC- & 1,979 \\
G/M prog & 1,881 \\
pDC & 1,758 \\
HSC & 1,703 \\
cDC2 & 1,702 \\
Lymph prog & 1,681 \\
Transitional B & 1,575 \\
Proerythroblast & 1,512 \\
CD8+ T CD57+ CD45RO+ & 1,470 \\
Normoblast & 1,435 \\
CD8+ T CD57+ CD45RA+ & 1,303 \\
CD8+ T TIGIT+ CD45RO+ & 1,160 \\
CD4+ T activated integrinB7+ & 1,056 \\
CD8+ T TIGIT+ CD45RA+ & 1,032 \\
CD8+ T CD49f+ & 912 \\
CD8+ T CD69+ CD45RO+ & 897 \\
B1 B IGKC+ & 820 \\
MAIT & 756 \\
CD8+ T CD69+ CD45RA+ & 740 \\
MK/E prog & 690 \\
gdT CD158b+ & 674 \\
B1 B IGKC- & 613 \\
T reg & 609 \\
ILC1 & 552 \\
Plasma cell IGKC+ & 288 \\
Plasma cell IGKC- & 239 \\
ILC & 238 \\
Plasmablast IGKC+ & 232 \\
gdT TCRVD2+$^\dagger$ & 191 \\
Plasmablast IGKC-$^\dagger$ & 130 \\
CD4+ T CD314+ CD45RA+$^\dagger$ & 93 \\
dnT$^\dagger$ & 56 \\
CD8+ T naive CD127+ CD26- CD101-$^\dagger$ & 42 \\
T prog cycling$^\dagger$ & 24 \\
cDC1$^\dagger$ & 18 \\
\midrule
Total & 90,261 \\
\bottomrule
\end{tabular}
\end{table}

\begin{table}[p]
\centering
\caption{Multiome: nuclei per cell type and batch (site--donor). A dash marks a type absent from a batch. The three site-4 batches form the cell-level test set.}
\label{tab:multiome_by_batch}
\scriptsize
\setlength{\tabcolsep}{3pt}
\begin{tabular}{lrrrrrrrrrrrrr}
\toprule
Cell type & s1d1 & s1d2 & s1d3 & s2d1 & s2d4 & s2d5 & s3d3 & s3d6 & s3d7 & s3d10 & s4d1 & s4d8 & s4d9 \\
\midrule
CD8+ T & 725 & 1,632 & 1,020 & 169 & 230 & 752 & 378 & 488 & 110 & 432 & 759 & 4,089 & 805 \\
CD14+ Mono & 921 & 704 & 242 & 1,308 & 1,140 & 1,716 & 855 & 208 & 674 & 1,603 & 1,014 & 458 & -- \\
NK & 591 & 982 & 696 & 269 & 136 & 848 & 477 & 126 & 256 & 545 & 1,124 & 421 & 458 \\
CD4+ T activated & 836 & 709 & 392 & 446 & 385 & 493 & 228 & 276 & 102 & 207 & 619 & 684 & 149 \\
Naive CD20+ B & 386 & 121 & 133 & 289 & 590 & 105 & 393 & 18 & 190 & 421 & 626 & 883 & 897 \\
Erythroblast & 797 & 136 & 374 & 478 & 1,485 & 64 & 246 & 84 & 31 & 391 & 474 & 266 & 90 \\
CD4+ T naive & 543 & 1,289 & 533 & 254 & 214 & 163 & 319 & 58 & 98 & 106 & 401 & 299 & 121 \\
Transitional B & 83 & 40 & 51 & 31 & 185 & -- & 146 & 18 & 30 & 359 & 695 & 630 & 542 \\
Proerythroblast & 65 & 30 & 116 & 85 & 316 & 18 & 197 & 27 & 21 & 111 & 322 & 800 & 192 \\
CD16+ Mono & 66 & 179 & 26 & 209 & 177 & 380 & 41 & 18 & 41 & 253 & 251 & 99 & 154 \\
B1 B & 201 & 127 & 73 & 67 & -- & 123 & 217 & 55 & 56 & 127 & 420 & 326 & 98 \\
Normoblast & 387 & 69 & 90 & 146 & 641 & 17 & 57 & 18 & 9 & 129 & 97 & 90 & 30 \\
Lymph prog & 52 & 28 & 57 & 109 & 286 & -- & 77 & 15 & 17 & 231 & 291 & 327 & 289 \\
G/M prog & 110 & 27 & -- & 60 & 92 & -- & 32 & 119 & 8 & 432 & 171 & 66 & 86 \\
pDC & 92 & 53 & 31 & 89 & 48 & 55 & 102 & 39 & 55 & 427 & 157 & 24 & 19 \\
HSC & 41 & 33 & 24 & 50 & 100 & 20 & 122 & 38 & 34 & 291 & 139 & 108 & 72 \\
CD8+ T naive & -- & -- & 199 & -- & -- & -- & 228 & -- & -- & 366 & -- & -- & 219 \\
MK/E prog & 64 & 71 & 24 & 33 & 33 & -- & 71 & 48 & 14 & 125 & 201 & 120 & 80 \\
cDC2 & 36 & 89 & 74 & 34 & 53 & 30 & 114 & 26 & 25 & 83 & 111 & 160 & 24 \\
ILC & 102 & 367 & 124 & 18 & -- & 95 & -- & -- & -- & 129 & -- & -- & -- \\
Plasma cell & 59 & 13 & -- & 76 & -- & 16 & 25 & -- & -- & 13 & 151 & 26 & -- \\
ID2-hi myeloid prog & 67 & 41 & -- & -- & -- & -- & -- & -- & -- & -- & -- & -- & -- \\
\midrule
Total & 6,224 & 6,740 & 4,279 & 4,220 & 6,111 & 4,895 & 4,325 & 1,679 & 1,771 & 6,781 & 8,023 & 9,876 & 4,325 \\
\bottomrule
\end{tabular}
\end{table}

\begin{table}[p]
\centering
\caption{CITE-seq: cells per cell type and batch (site--donor). A dash marks a type absent from a batch.}
\label{tab:cite_by_batch}
\scriptsize
\setlength{\tabcolsep}{3pt}
\begin{tabular}{lrrrrrrrrrrrr}
\toprule
Cell type & s1d1 & s1d2 & s1d3 & s2d1 & s2d4 & s2d5 & s3d1 & s3d6 & s3d7 & s4d1 & s4d8 & s4d9 \\
\midrule
CD14+ Mono & 734 & 726 & 309 & 2,958 & 1,294 & 2,561 & 2,915 & 2,927 & 6,056 & 637 & 466 & 110 \\
CD4+ T activated & 381 & 848 & 547 & 690 & 364 & 1,009 & 816 & 544 & 380 & 577 & 134 & 676 \\
CD4+ T naive & 591 & 304 & 833 & 883 & 266 & 631 & 681 & 349 & 219 & 583 & 87 & 470 \\
NK & 94 & 458 & 271 & 282 & 76 & 1,128 & 84 & 415 & 1,242 & 511 & 268 & 605 \\
Reticulocyte & 605 & 279 & 929 & 270 & 580 & 116 & 587 & 244 & 291 & 194 & 50 & 127 \\
Erythroblast & 228 & 57 & 228 & 1,202 & 471 & 61 & 622 & 524 & 307 & 150 & 35 & 154 \\
Naive CD20+ B IGKC+ & 454 & 150 & 265 & 585 & 629 & 129 & 318 & 105 & 88 & 306 & 243 & 718 \\
CD8+ T naive & 229 & 129 & 878 & 312 & 89 & 73 & 349 & 157 & 116 & 179 & 193 & 403 \\
CD16+ Mono & 45 & 142 & 28 & 271 & 129 & 692 & 468 & 248 & 353 & 81 & 31 & 147 \\
NK CD158e1+ & 119 & 76 & -- & 259 & 80 & 740 & 135 & 452 & -- & 163 & -- & 143 \\
Naive CD20+ B IGKC- & 316 & 112 & 171 & 394 & -- & 89 & 99 & 86 & 71 & 136 & 133 & 372 \\
G/M prog & 10 & 86 & 45 & 56 & 308 & 24 & 299 & 452 & 237 & 152 & 47 & 165 \\
pDC & 57 & 67 & 20 & 332 & 63 & 165 & 261 & 337 & 364 & 46 & 14 & 32 \\
HSC & 58 & 72 & 71 & 145 & 193 & 38 & 162 & 400 & 178 & 90 & 86 & 210 \\
cDC2 & 35 & 62 & 54 & 166 & 127 & 102 & 247 & 396 & 314 & 98 & 60 & 41 \\
Lymph prog & 67 & 27 & 87 & 90 & 208 & 5 & 144 & 236 & 154 & 148 & 160 & 355 \\
Transitional B & 76 & 90 & 250 & 43 & 150 & 9 & 76 & 205 & 119 & 79 & 130 & 348 \\
Proerythroblast & 100 & 10 & 26 & 92 & 44 & -- & 135 & 259 & 225 & 108 & 172 & 341 \\
CD8+ T CD57+ CD45RO+ & -- & 119 & -- & 48 & 15 & 54 & 33 & 110 & 52 & 38 & 744 & 257 \\
Normoblast & 266 & 56 & 129 & 189 & 171 & 17 & 257 & 164 & 40 & 62 & 23 & 61 \\
CD8+ T CD57+ CD45RA+ & 33 & 136 & 251 & 30 & -- & 36 & -- & 457 & -- & 53 & 258 & 49 \\
CD8+ T TIGIT+ CD45RO+ & 35 & 52 & 37 & 118 & 26 & 486 & 12 & 205 & -- & 102 & -- & 87 \\
CD4+ T activated integrinB7+ & 124 & 123 & 141 & 105 & 29 & 78 & 97 & 128 & -- & 91 & 49 & 91 \\
CD8+ T TIGIT+ CD45RA+ & 58 & -- & 25 & -- & -- & 433 & 51 & 244 & 57 & -- & 74 & 90 \\
CD8+ T CD49f+ & 58 & 48 & 49 & 64 & 33 & 123 & 48 & 226 & 15 & 121 & 40 & 87 \\
CD8+ T CD69+ CD45RO+ & 55 & 8 & 32 & 88 & -- & -- & 31 & 285 & 50 & 122 & 51 & 175 \\
B1 B IGKC+ & 57 & 42 & 41 & 86 & -- & 87 & 118 & 89 & 84 & 67 & 72 & 77 \\
MAIT & 28 & 107 & 109 & 34 & 6 & 27 & 23 & 60 & 94 & 40 & 71 & 157 \\
CD8+ T CD69+ CD45RA+ & 87 & 20 & 33 & 92 & 21 & -- & 14 & 240 & -- & 85 & 110 & 38 \\
MK/E prog & 14 & 9 & 8 & 43 & 80 & 10 & 81 & 168 & 81 & 31 & 27 & 138 \\
gdT CD158b+ & -- & 158 & 62 & -- & -- & -- & -- & -- & 22 & 37 & -- & 395 \\
B1 B IGKC- & 42 & 52 & 48 & 49 & -- & 52 & 89 & 67 & 47 & 59 & 46 & 62 \\
T reg & 88 & 21 & 36 & 196 & 37 & -- & 50 & -- & -- & 95 & 26 & 60 \\
ILC1 & 30 & 54 & 40 & 24 & 14 & 96 & 30 & 82 & 35 & 30 & 18 & 99 \\
Plasma cell IGKC+ & 9 & 6 & 9 & 53 & 15 & 10 & 60 & 43 & 41 & 15 & 7 & 20 \\
Plasma cell IGKC- & 3 & 2 & -- & 81 & -- & 3 & 38 & 29 & 64 & 10 & 4 & 5 \\
ILC & -- & 86 & -- & 48 & -- & 1 & -- & -- & -- & 103 & -- & -- \\
Plasmablast IGKC+ & 18 & 2 & -- & 45 & -- & 19 & 42 & 68 & 22 & 16 & -- & -- \\
gdT TCRVD2+ & 16 & 44 & 10 & 33 & 56 & -- & 12 & -- & -- & 20 & -- & -- \\
Plasmablast IGKC- & 7 & 4 & -- & 9 & -- & 18 & 37 & 34 & -- & 21 & -- & -- \\
CD4+ T CD314+ CD45RA+ & -- & 92 & 1 & -- & -- & -- & -- & -- & -- & -- & -- & -- \\
dnT & -- & -- & 33 & -- & 10 & -- & -- & -- & 13 & -- & -- & -- \\
CD8+ T naive CD127+ CD26- CD101- & -- & 42 & -- & -- & -- & -- & -- & -- & -- & -- & -- & -- \\
T prog cycling & -- & -- & -- & -- & -- & -- & -- & -- & 24 & -- & -- & -- \\
cDC1 & -- & -- & -- & -- & -- & -- & -- & -- & 18 & -- & -- & -- \\
\midrule
Total & 5,227 & 4,978 & 6,106 & 10,465 & 5,584 & 9,122 & 9,521 & 11,035 & 11,473 & 5,456 & 3,929 & 7,365 \\
\bottomrule
\end{tabular}
\end{table}

\clearpage
\section{Modeling and experiments}

\subsection{Embeddings and encoder initialization}
\label{supp:embeddings}
 All single-modality representations come from Liam \citep{liam, liam_reproducibility}, a batch-adversarial variational autoencoder, trained on one modality at a time with no cell-type labels and no access to the pairing; the model registers no label field, so none could enter training. At the cell-type level on the multiome data, the embeddings are our own training of Liam's single-modality recipes (\texttt{rna\_only}, \texttt{atac\_only}, $20$ latent dimensions) on all $69,249$ nuclei, five training seeds ($0$, $3$, $11$, $15$, $19$), one per dataset seed; we verified that they separate cell types no better than the authors' released label-free runs (k-NN balanced accuracy $0.778$ and $0.722$ for RNA and ATAC, against $0.797$ and $0.731$). On CITE-seq we use the authors' released single-modality embeddings (\texttt{rna\_only}, \texttt{adt\_only}, seeds $0$, $71$, $236$, $415$, $994$; for $81,241$ cells). At the cell level, the encoders are initialized from the authors' trained single-modality models for each dataset (seeds $0$, $994$, $236$, $71$, $415$ for dataset seeds $0$ to $4$), which were trained on the $62,501$ cells of the competition's phase-2 file, test nuclei included, one modality at a time; our encoders take the same inputs those models saw, log-transformed gene counts, binarized peak accessibility, and centered-log-ratio protein counts. The two Liam latent spaces are unaligned: the cosine similarity between the RNA and ATAC embeddings of the same cell is 0.43, against 0.42 for two random cells. At the cell level, the first layer of each encoder holds most of the parameters (the ATAC leg's first layer alone has $116{,}490 \times 128$ weights), and a posterior over it dominated the KL term without changing the posterior over the remaining layers, so it is kept as a point estimate. A head that concatenates the two representations, as at the cell-type level, is not usable at the cell level: scoring every pair of a block of $n$ cells would take $n^2$ forward passes through the head, whereas the cosine score of every pair is one matrix product.
 
\subsection{Loss function and training}
\label{supp:training}
\textbf{Cell-type level.} The loss is the binary cross-entropy of $p$ against the pair label plus the Kullback--Leibler divergence of the posterior from the prior, divided by the number of training pairs, so that the KL term is weighted as in the evidence lower bound. We train with Adam \citep{kingma2015adam} at learning rate $2 \times 10^{-3}$, the variational scale parameters at their own learning rate of $5 \times 10^{-2}$, in batches of 512 pairs, for at most 200 epochs, evaluating the validation loss of the posterior predictive (the mean probability over 20 draws; 200 draws are used only at test time) every 25 optimizer steps and stopping after 40 evaluations with no improvement. No class weights are needed because every dataset is balanced by construction (Section~\ref{sec:experiments}).

\textbf{Cell level.} Training proceeds on blocks of $1,024$ cells, each cell contributing its GEX and its ATAC view. The loss is the symmetric InfoNCE \citep{oord2018representation, radford2021learning} over the block's $1{,}024 \times 1{,}024$ score matrix,
\[
\mathcal{L} = \frac{1}{2n} \sum_{i=1}^{n} \left[ -\log \frac{e^{S_{ii}}}{\sum_{j} e^{S_{ij}}} - \log \frac{e^{S_{ii}}}{\sum_{j} e^{S_{ji}}} \right],
\]
so that each cell's own partner is the positive and the other 1,023 cells of the block are its negatives, in both directions, plus the KL term divided by the number of training cells. Because the negatives come from the same batch as the query, the model is never rewarded for telling batches apart. Training views are augmented by binomial thinning: each GEX count is kept with probability 0.8, and each nonzero ATAC entry is kept with probability 0.8, so that the two views of one cell differ between epochs in which reads were observed but not in what the cell is. Thinning is not applied to ADT, whose values are not counts. We train with Adam at learning rate $10^{-3}$, the pretrained encoders at a tenth of it, and stop on the InfoNCE loss of 16 fixed validation blocks, the same blocks for every arm. No cell-type label is read at any point of training or model selection.

\subsection{Uncertainty scores and controls}
\label{supp:scores}
\textbf{Cell-type-level prediction and uncertainty score.} At test time the inference on each pair is repeated for $S = 200$ posterior draws (i.e., Monte-Carlo sampling), giving a distribution of probabilities per pair. Its mean $\bar p$ is the prediction, called positive at $\bar p \geq 0.5$, and its standard deviation $p_{\mathrm{sd}}$ is the uncertainty score. The deterministic control for every reading of $p_{\mathrm{sd}}$ is the confidence margin, $0.5 - |\bar p - 0.5|$, which is high when the prediction is close to the decision boundary and is available to the deterministic network as well; whatever $p_{\mathrm{sd}}$ does that the margin does not is what the posterior adds.

\textbf{Cell-level uncertainty scores.} The posterior arm draws 200 weight samples per seed, each giving its own score matrix $S^{(s)}$; the draws of the five seeds are pooled, $S = 1{,}000$, and read on the ensemble assignment $\pi$. Two reads of the same draws give two uncertainty scores. The \emph{top-1 vote}
\[
u_i = \frac{1}{S} \sum_{s} \mathbf{1}\!\left[ \arg\max_j S^{(s)}_{ij} \neq \pi(i) \right]
\]
is the share of draws that would have matched $i$ to a cell other than its assigned partner. The \emph{transport-plan read}
\[
v_i = 1 - \frac{1}{S} \sum_{s} P^{(s)}_{i,\pi(i)}, \qquad P^{(s)} = \mathrm{Sinkhorn}\!\left(S^{(s)} / T_v\right),
\]
is one minus the mass the draws' entropic transport plans \citep{cuturi2013sinkhorn} place on the assigned partner; it reads the whole row of each draw rather than only its argmax. Its temperature $T_v \in \{1, 0.5, 0.3\}$ is chosen the same way as the control's below. Both are high when the posterior does not agree with the assignment.

\textbf{Cell-level deterministic controls.} The deterministic counterpart of $v_i$ is the transport weight of the mean score matrix, $1 - P_{i,\pi(i)}$ with $P = \mathrm{Sinkhorn}(\bar S / T)$: the same read of the same assignment with no posterior. Its temperature $T$ is chosen from nine values between 0.01 and 1 on the AUROC for any wrong assignment, never on the error type under test, and the temperature that would have been chosen on the target is reported beside it, so that the control is never tuned on the question it is asked. We also report the row margin, the gap between the assigned partner's score and the runner-up's, and seed stability, the share of the five seeds whose own Hungarian assignment disagrees with $\pi$, which is the uncertainty a deep ensemble of the deterministic arm would give.

\textbf{Cell-type level baselines.} Three baselines use the same frozen embeddings and the same pair labels with no network: $k$-nearest-neighbor label transfer in each modality ($k = 15$), with the probability that the two transferred labels agree as the score; an orthogonal Procrustes rotation fit on the training pairs, with the cosine similarity Platt-scaled to a probability; and a logistic regression on the concatenation and the outer product of the two embeddings. \chg\rev{Figures~\ref{fig:type_level} and \ref{fig:cite_both_jobs} also show an ensemble of the logistic pair classifier, labeled logistic (ens): five classifiers, each fitted on a bootstrap resample of the training pairs, with the members' mean as the prediction and the standard deviation of their probabilities as the uncertainty score, the same two numbers a deep ensemble gives.} 

\subsection{Data constructions}
\label{supp:constructions}

\paragraph{Cell-type level.}
All constructions start from the same pair table: for each cell type we draw $\min(N_t, 100)$ cells, split them 70/15/15 into train, validation, and test, and form all $m^2$ positive and $m^2$ negative pairs within each split, where $m$ is the number of drawn cells. Negative partners are drawn uniformly over the other types, not by prevalence, so rare types are represented among the negatives. Pairs are formed in two populations: without regard to batch (\emph{xbatch}, called cross-batch in the main text) and within a batch (\emph{wbatch}, both cells from the same site and donor); the main text reports xbatch unless stated. The two populations exist to check that the model does not learn the batch instead of the cell type: a model that used batch identity as a shortcut would score within-batch pairs and batch-unaware pairs differently, and on the multiome data the two are classified the same, AUROC 0.977 against 0.976 (Section~\ref{sec:results_type}), with the caveat that the input embeddings were made batch-adversarial by Liam before this model saw them, so part of that is inherited from the input.

\emph{Mix} trains on all types and all batches. \emph{LOCTO} removes one type from train and validation and adds it back only in the test pairs, once as left leg and once as right leg; we run all 22 multiome types and all 38 CITE-seq types. \emph{LOFO} removes every type of one lineage family (Table~\ref{tab:families}); the seven families give seven runs. \emph{LTFO} removes two families at a time. In each held-out construction the test set contains three pair kinds: held-out versus seen (the novelty pairs), held-out versus held-out (positives the model has never seen either side of), and seen versus seen (the control, which should behave as in mix). We report them separately.

\emph{Few-shot} restarts LOFO with $k \in \{0, 5, 10, 20, 30, 50\}$ cells of each type of the held-out family returned to training, drawn from a pool of 50 cells per type; the other 50 cells per type are the test pool, the same for every $k$. The $k$ cells are chosen by one of three rules: \emph{random} (uniform), \emph{hard} (the $k$ cells closest in the RNA embedding to a seen type), and \emph{diverse} (farthest-point sampling within the type).

\paragraph{Cell level.}
The cell-level matcher trains on the competition split of the multiome data. \emph{Challenge} holds out the 20{,}009 site-4 nuclei (batches s4d1, s4d8, s4d9) as the test set, and splits the remaining cells 85/15 within each type$\times$batch cell into train and validation. \emph{Challenge-LOCTO} additionally removes one of eight cell types from train and validation and scores the held-out cells against the seen cells in the same batch. Matching is always within batch, by the Hungarian assignment on the mean score matrix.

\paragraph{Seeds.}
Every construction is run with five seeds. At the cell-type level the seed fixes the Liam embedding (seeds 0, 3, 11, 15, 19 for multiome; 0, 71, 236, 415, 994 for CITE-seq), the cell draw, and the split. At the cell level the seed fixes the Liam initialization of the encoders (0, 994, 236, 71, 415) and the training order. All arms within one seed see the same pairs.

Table~\ref{tab:experiments} lists every construction with its arms and its number of runs.
\begin{table}[htbp]
\centering
\caption{The experiments. Each cell-type-level run is one trained model on one dataset (construction $\times$ seed $\times$ pair population); the arms share every dataset. Each cell-level run is one trained matcher.}
\label{tab:experiments}
\resizebox{\textwidth}{!}{\begin{tabular}{llllr}
\toprule
level & data & construction & arms & runs per arm \\
\midrule
cell type & multiome & mix & BNN, legs, MCD, ENS, DET, $k$-NN, cosine, logistic & 10 \\
cell type & multiome & LOCTO (22 types) & same & 220 \\
cell type & multiome & LOFO (7 families) & BNN & 70 \\
cell type & multiome & LTFO (21 pairs) & BNN & 210 \\
cell type & multiome & few-shot ($k \in \{0, 5, 10, 20, 30, 50\}$, 3 rules) & BNN & 1{,}260 \\
cell type & multiome & width ladder ($\lambda \in \{0, 0.1, 0.3\}$) & BNN & 660 \\
cell type & CITE-seq & mix + LOCTO (38 types) & BNN, MCD, ENS, DET, $k$-NN, cosine, logistic & 390 \\
cell & multiome & challenge & BNN, DET & 5 \\
cell & multiome & challenge-LOCTO (8 types) & BNN, DET & 40 \\
\bottomrule
\end{tabular}}
\end{table}

\subsection{Metrics}
\label{supp:metrics}
\textbf{Measures.} Predictive performance is the classification AUROC and accuracy of $\bar p$ at the cell-type level and the matching accuracy at the cell level; at the cell level we also report the accuracy as a function of the number of seeds averaged, so that the reader sees what the ensemble adds. \emph{Error detection} is the AUROC of an uncertainty score for separating wrong predictions from correct ones; at the cell level, wrong matches are further split by whether the assigned partner is of the query's own type, a lineage relative, or another family, and each kind is read separately, because a wrong match to the same type is close to harmless for anything downstream that works at the type level, and a wrong match across families is the harmful kind. \emph{Abstention} ranks the predictions by the score and discards the least certain share; we report the accuracy on what is kept and the share of the errors removed, against a blind rule that discards at random. \emph{Novelty} is, within one held-out run, the AUROC of the score for separating the held-out type's cells from seen cells; at the cell-type level it is read on positive pairs, since negative pairs carry almost no uncertainty on either side. \emph{Error structure}, at the cell level, is the share of wrong matches whose partner is of the query's own type and, of those that leave the type, the share whose partner is a lineage relative, each against the same share with the partners permuted within the batch.

\textbf{Statistics.} At the cell-type level the run is the unit of replication: every statistic is computed within a run and reported as mean and standard deviation over runs, and LOCTO statistics are clustered by held-out type, with a bootstrap over types for the interval. At the cell level every comparison of two scores is made within one test batch and reported per batch, with a bootstrap interval over queries. Every reading was written down with its prediction and its bar before the run, and the readings that missed their bar are reported as such; the record is in the repository.

\subsection{The width of the posterior}
\label{supp:kl}

The mean-field posterior is trained with the KL term of the ELBO scaled by a factor $\lambda$; $\lambda = 1$ is the standard ELBO and the setting used everywhere in the main text. With $\lambda = 0$ the posterior collapses onto its mean, and at $\lambda = 140$ (the KL divided by the number of training cells instead of the number of training pairs; 9{,}800 pairs over 70 cells per type) the classifier breaks (mix AUROC 0.849 $\pm$ 0.127). We ran the region between: $\lambda \in \{0, 0.1, 0.3, 1\}$ on every LOCTO type and both pair populations, 220 runs per width, plus $\lambda = 0.01$ on the mix datasets only. The posterior width $\sigma$ is the mean standard deviation of the variational weights against a prior of 1.0. Predictions were recorded before the run.

Table~\ref{tab:kl} gives, per width and clustered by held-out type ($n = 22$), the three uncertainty measures of the main text against their control: error detection of $p_{\mathrm{sd}}$ minus that of the confidence margin, the gain in accuracy from keeping the most certain half by $p_{\mathrm{sd}}$ minus the same gain by the margin, and novelty detection by $p_{\mathrm{sd}}$ (held-out positives against seen positives).

\begin{table}[htbp]
\centering
\caption{The posterior's width against what it buys. Test split, clustered by held-out type, $n = 22$ types; brackets are 95\% intervals. Classification AUROC is on the mix datasets, $n = 10$ runs. The last column is the per-pair Spearman correlation of $p_{\mathrm{sd}}$ with the confidence margin, mean $\pm$ SD over the mix runs.}
\label{tab:kl}
\small
\resizebox{\textwidth}{!}{\begin{tabular}{lcccccc}
\toprule
$\lambda$ & $\sigma$ & mix AUROC & error AUROC, $p_{\mathrm{sd}}$ $-$ margin & gain at 50\%, $-$ margin & novelty AUROC, $p_{\mathrm{sd}}$ & Spearman($p_{\mathrm{sd}}$, margin) \\
\midrule
0    & 0.004 & 0.978 & $-0.001$ [$-0.002$, $+0.001$] & $+0.001$ & 0.584 & 0.956 $\pm$ 0.027 \\
0.01 & 0.042 & 0.978 & --                            & --       & --    & 0.984 $\pm$ 0.011 \\
0.1  & 0.262 & 0.976 & $+0.005$ [$+0.003$, $+0.007$] & $+0.003$ & 0.612 & 0.989 $\pm$ 0.008 \\
0.3  & 0.489 & 0.976 & $+0.007$ [$+0.004$, $+0.010$] & $+0.005$ & 0.623 & 0.987 $\pm$ 0.008 \\
1    & 0.750 & 0.976 & $+0.007$ [$+0.003$, $+0.011$] & $+0.006$ & 0.627 & 0.986 $\pm$ 0.005 \\
\bottomrule
\end{tabular}}
\end{table}

\chg\rev{Three readings of the table.} Classification does not depend on the width: across the 200-fold range from $\sigma = 0.004$ to $0.750$ the mix AUROC moves by less than the seed-to-seed spread inside one width. Every uncertainty measure is monotone in the width, and the collapsed posterior is the worst of the four on all three; novelty detection loses 0.043 when the posterior has no width, the largest effect any $\lambda$ produces. The effect saturates: $\lambda = 0.3$ and $\lambda = 1$ agree on error detection and lie within 0.005 on novelty detection, so nothing changes above $\sigma \approx 0.5$. The standard ELBO weight is the best of the region we sampled, and we did not tune it. The width does not change the relationship between $p_{\mathrm{sd}}$ and the margin: their per-pair correlation is 0.96 to 0.99 at every width, which is what Section~\ref{sec:results_type} states. Width sets how much the posterior carries; where it sits in the network sets what it carries.

\section{Additional results}
\label{supp:additional}

This section holds the tables and figures the main text cites at the cell-type level and the cell level. Every number in them is from the same runs as the main text: \chg\rev{230 runs at the cell-type level, 115 per pair population (five mix, 110 LOCTO)}, and the 20{,}009 test cells of the competition split at the cell level. The full per-population tables, the 44 per-type uncertainty distributions, and the readings that missed their bar are in the repository, not here.

\begin{figure}[htbp]\centering
\includegraphics[width=\textwidth]{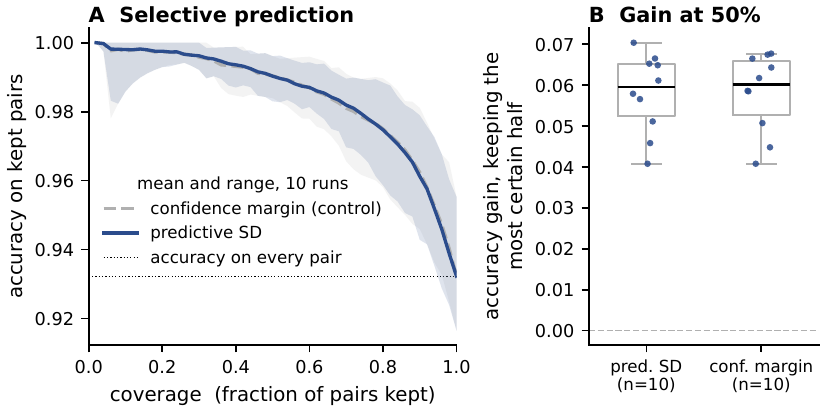}
\caption{Selective prediction on the mix datasets, ten runs. (A) Accuracy on the pairs kept when the least certain are discarded, ranked by predictive SD (solid) or by the confidence margin (dashed); the band is the run-to-run range and the dotted line is accuracy on every pair. (B) Accuracy gain from keeping the most certain half, one point per run.}
\label{fig:risk_coverage}\end{figure}

\begin{figure}[htbp]\centering
\includegraphics[width=\textwidth]{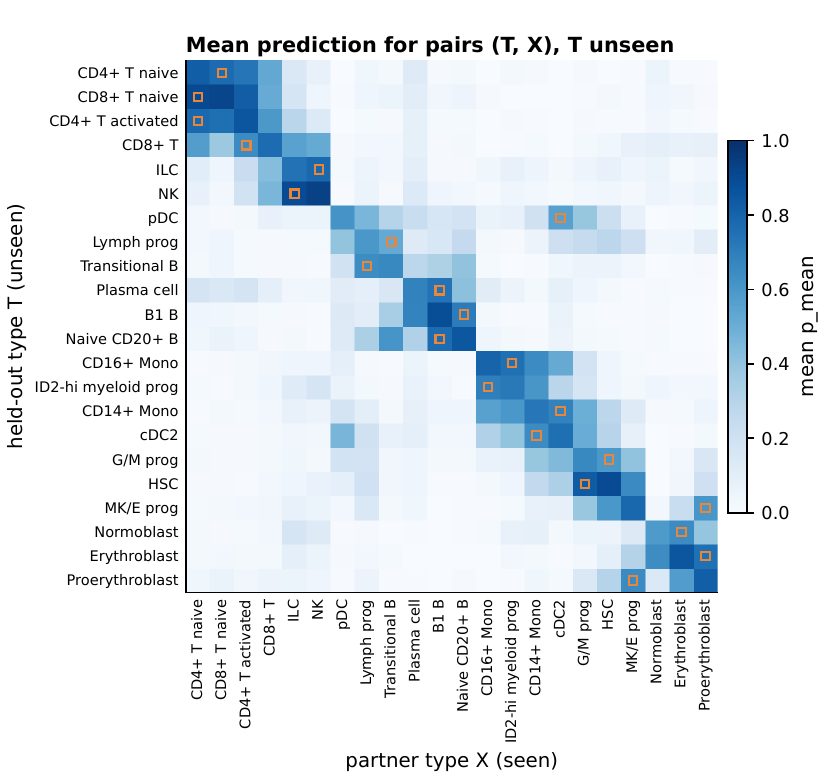}
\caption{Learned similarity between cell types. Entry $(T, X)$ is the mean predicted probability that a cell of type $T$ and a cell of type $X$ share a type, read from the LOCTO runs in which $T$ was held out (five seeds, both pair populations); the diagonal is $T$'s own positives. Rows and columns are ordered by similarity. A square marks each row's largest off-diagonal entry.}
\label{fig:similarity}\end{figure}

\begin{figure}[htbp]\centering
\includegraphics[width=\textwidth]{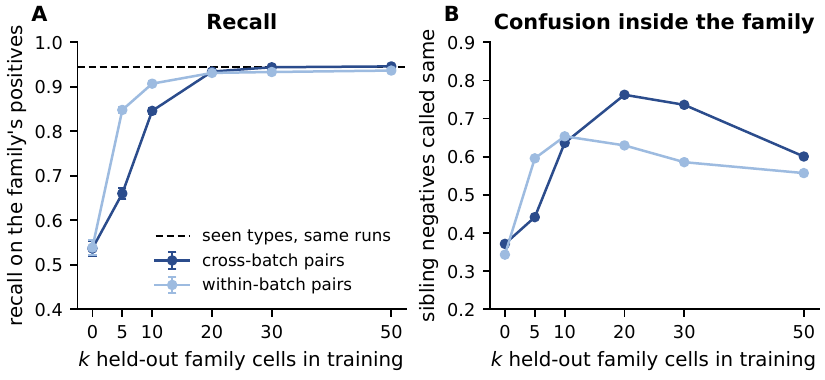}
\caption{Few-shot recovery of a held-out lineage family. (A) Recall on the family's positives against the number $k$ of its cells returned to training, on both pair populations, with the seen-type recall of the same runs as reference; error bars are the standard error over \chg\rev{330 per-type rows per point (22 types $\times$ five seeds $\times$ three rules)}. (B) The share of the family's sibling negatives called same, against $k$.}
\label{fig:fewshot}\end{figure}

\paragraph{Cell-type level.}
Table~\ref{tab:methods} is the comparison on seen types: accuracy, AUROC, Brier score, NLL, and ECE of the seven methods on the mix datasets, cross-batch pairs. The four network arms are within 0.006 accuracy of each other and ahead of the three baselines on every proper scoring rule; $k$-NN label transfer is the outlier, with a Brier score of 0.134 because its score is a vote agreement and not a calibrated probability. Table~\ref{tab:bothjobs} is the two-jobs reading of the same methods on the 110 LOCTO runs of each pair population, the numbers behind Figure~\ref{fig:type_level}A--B: the two AUROCs per method and the share of runs in which one score clears 0.55 on both. Figure~\ref{fig:risk_coverage} is selective prediction on the mix datasets: accuracy on the kept pairs against coverage when pairs are ranked by $p_{\mathrm{sd}}$ and by the confidence margin, and the gain at half coverage per run. The two curves lie on each other, which is the tie stated in Section~\ref{sec:results_type}. Figure~\ref{fig:similarity} is the $22 \times 22$ learned similarity matrix assembled from the LOCTO runs, ordered by similarity so that the lineage blocks are visible: T cells, ILC and NK, pDC, the B lineage, monocytes and dendritic cells, progenitors, and the erythroid chain. Figure~\ref{fig:fewshot} is the few-shot recovery of a held-out family: recall on the family's positives against the number of its cells returned to training, and the share of sibling negatives called same, which rises up to $k = 20$ and then turns.

\begin{figure}[htbp]\centering
\includegraphics[width=\textwidth]{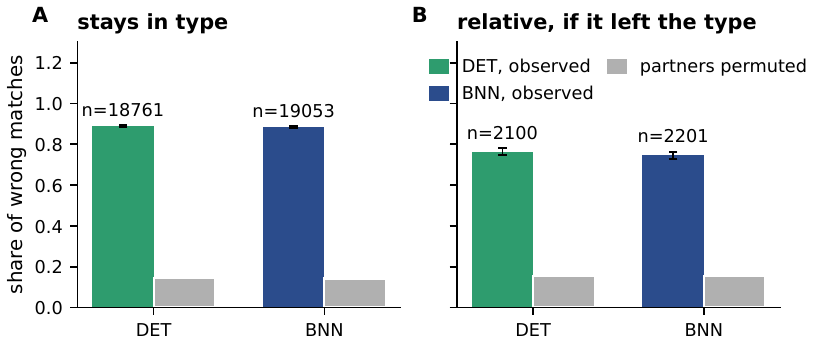}
\caption{Where a wrong assignment goes, both arms. (a) The share of wrong assignments whose partner is another cell of the query's own type. (b) Of those that leave the type, the share whose partner is a lineage relative. Grey bars are the same shares with the partners permuted within batch; error bars are bootstrap intervals over queries.}
\label{fig:errors_where}\end{figure}

\begin{figure}[htbp]\centering
\includegraphics[width=\textwidth]{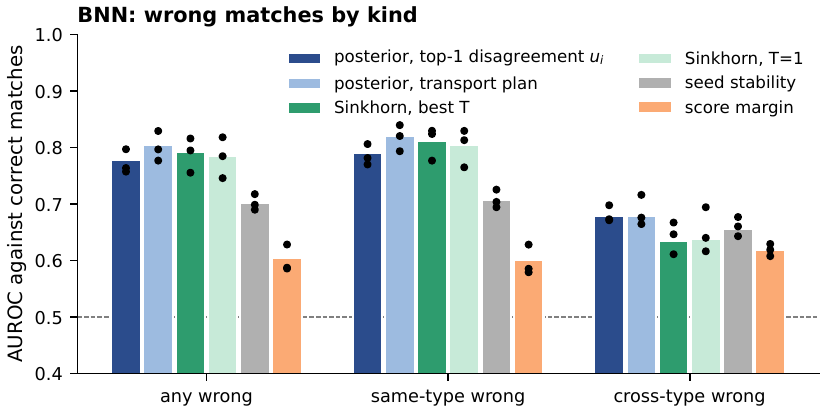}
\caption{Detecting a wrong assignment, by kind of error, Bayesian arm. AUROC of each score for separating wrong assignments of the given kind from correct matches, pooled over the three test batches (bars) and per batch (points).}
\label{fig:errors_harmful}\end{figure}

\begin{figure}[htbp]\centering
\includegraphics[width=\textwidth]{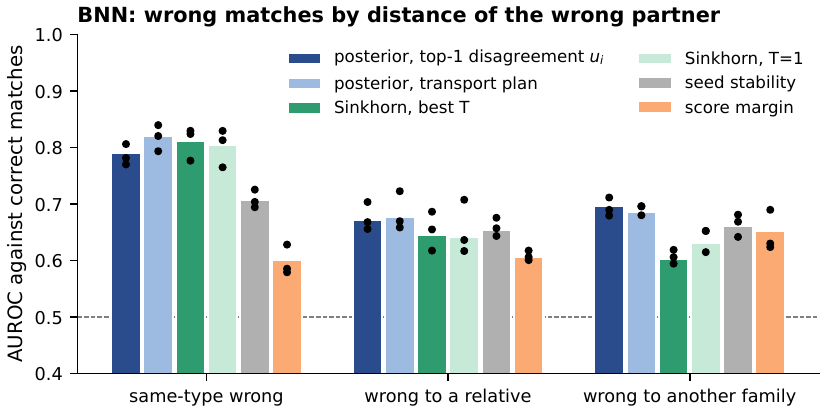}
\caption{The same scores on the cross-type errors, split by whether the wrong partner is a lineage relative of the query's type or belongs to another family.}
\label{fig:errors_distance}\end{figure}

\begin{figure}[htbp]\centering
\includegraphics[width=0.9\textwidth]{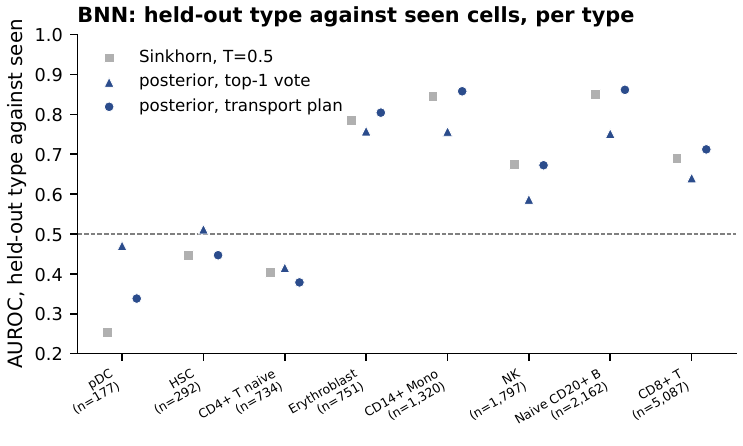}
\caption{A type never seen in training, at the cell level, Bayesian arm, 50 posterior draws per seed. For each of eight held-out types, the AUROC of each score for separating the held-out cells from seen cells in the same test batches; types are ordered by the number of held-out test cells.}
\label{fig:locto_cell}\end{figure}

\paragraph{Cell level.}
Figure~\ref{fig:errors_where} is where a wrong assignment goes, in both arms, against the permuted control: the share that stays in the query's own type and, of those that leave it, the share that goes to a lineage relative. Figure~\ref{fig:errors_harmful} is the detection of a wrong assignment by kind of error, the figure behind Table~\ref{tab:harmful}, with the per-batch points that the pooled bars hide. Figure~\ref{fig:errors_distance} splits the cross-type errors further by whether the wrong partner is a relative or belongs to another family; the posterior's lead over the Sinkhorn control is small on relatives and large on other families. Figure~\ref{fig:locto_cell} is the cell-level version of the held-out type: for each of eight held-out types, ordered by the number of held-out test cells, the AUROC of the Sinkhorn control and of the two posterior reads for separating held-out from seen cells. The three small types sit \chg\rev{at or} below chance for every score, which is the type-size dependence stated in Section~\ref{sec:results_cell}.

\begin{table}[htbp]
\centering
\caption{Classification and probability quality of the seven methods on pairs of seen types (mix, cross-batch pairs, test split, mean $\pm$ SD over five seeds). Every method reads the same frozen embeddings and the same pairs. Brier score, NLL, and ECE are lower-is-better; the three baselines use no network.}
\label{tab:methods}
\resizebox{\textwidth}{!}{\begin{tabular}{lccccc}
\toprule
method & accuracy & AUROC & Brier & NLL & ECE \\
\midrule
BASILISC (MFVI) & 0.932 $\pm$ 0.006 & 0.976 $\pm$ 0.002 & 0.053 $\pm$ 0.004 & 0.191 $\pm$ 0.012 & 0.014 $\pm$ 0.004 \\
MC dropout & 0.938 $\pm$ 0.003 & 0.981 $\pm$ 0.001 & 0.048 $\pm$ 0.002 & 0.168 $\pm$ 0.007 & 0.017 $\pm$ 0.004 \\
deep ensemble (5) & 0.938 $\pm$ 0.004 & 0.980 $\pm$ 0.002 & 0.048 $\pm$ 0.003 & 0.171 $\pm$ 0.012 & 0.011 $\pm$ 0.001 \\
deterministic (no sampling) & 0.932 $\pm$ 0.002 & 0.976 $\pm$ 0.002 & 0.052 $\pm$ 0.001 & 0.186 $\pm$ 0.007 & 0.013 $\pm$ 0.002 \\
linear pair classifier & 0.917 $\pm$ 0.006 & 0.969 $\pm$ 0.002 & 0.063 $\pm$ 0.003 & 0.218 $\pm$ 0.010 & 0.022 $\pm$ 0.005 \\
aligned cosine (Procrustes + Platt) & 0.914 $\pm$ 0.004 & 0.965 $\pm$ 0.003 & 0.066 $\pm$ 0.004 & 0.229 $\pm$ 0.011 & 0.013 $\pm$ 0.003 \\
$k$-NN label transfer & 0.792 $\pm$ 0.008 & 0.978 $\pm$ 0.002 & 0.134 $\pm$ 0.005 & 0.663 $\pm$ 0.089 & 0.199 $\pm$ 0.005 \\
\bottomrule
\end{tabular}}
\end{table}

\begin{table}[htbp]
\centering
\caption{One score, two jobs. For each method, the AUROC of its own uncertainty score for detecting its own errors on seen-type pairs and for detecting the held-out type's positives among seen positives, over the 110 LOCTO runs of each pair population, and the share of those runs in which the same score clears 0.55 on both. The deterministic network is the identical architecture with sampling switched off.}
\label{tab:bothjobs}
\resizebox{\textwidth}{!}{\begin{tabular}{lcccccc}
\toprule
 & \multicolumn{3}{c}{cross-batch pairs} & \multicolumn{3}{c}{within-batch pairs} \\
\cmidrule(lr){2-4}\cmidrule(lr){5-7}
method & error AUROC & unseen AUROC & both $>0.55$ & error AUROC & unseen AUROC & both $>0.55$ \\
\midrule
BASILISC (MFVI) & 0.834 & 0.619 & 69\% & 0.833 & 0.635 & 75\% \\
MC dropout & 0.843 & 0.707 & 88\% & 0.864 & 0.710 & 85\% \\
deep ensemble (5) & 0.850 & 0.671 & 82\% & 0.870 & 0.679 & 84\% \\
deterministic (no sampling) & 0.840 & 0.564 & 53\% & 0.856 & 0.582 & 61\% \\
linear pair classifier & 0.844 & 0.490 & 45\% & 0.852 & 0.451 & 38\% \\
aligned cosine (Procrustes + Platt) & 0.538 & 0.667 & 10\% & 0.532 & 0.695 & 15\% \\
$k$-NN label transfer & 0.812 & 0.607 & 54\% & 0.821 & 0.627 & 63\% \\
\bottomrule
\end{tabular}}
\end{table}


\section{Replication on a second modality pair: CITE-seq}
\label{supp:cite}

\begin{figure}[htbp]\centering
\includegraphics[width=\textwidth]{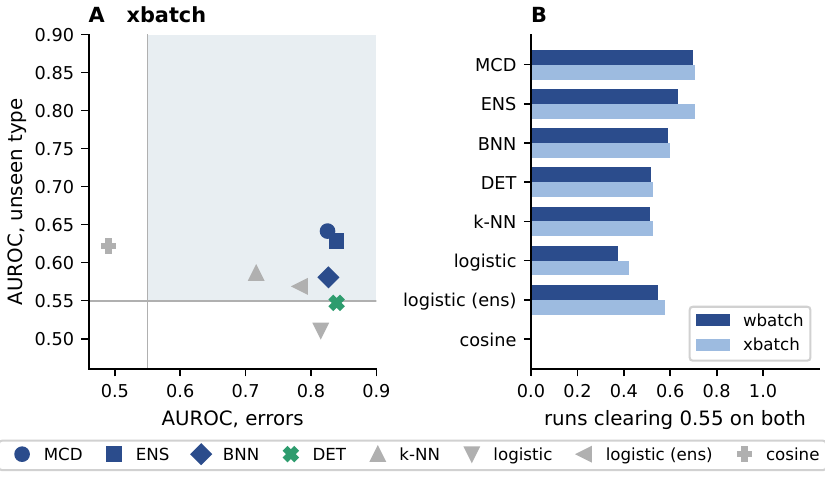}
\caption{CITE-seq: one score, two jobs, over 190 LOCTO runs per pair population. The aligned cosine clears 0.55 on both jobs in no run.}
\label{fig:cite_both_jobs}\end{figure}

\begin{figure}[htbp]\centering
\includegraphics[width=0.7\textwidth]{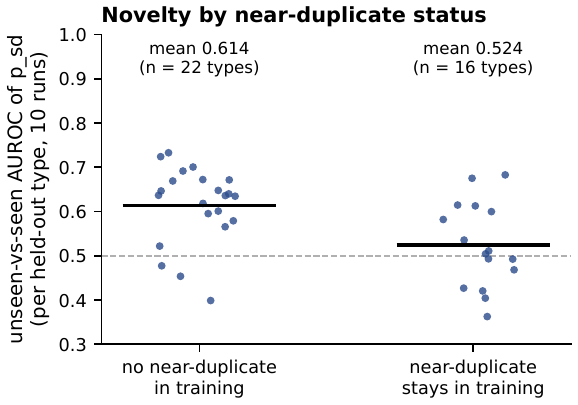}
\caption{CITE-seq: novelty detection per held-out type, split by whether a near-duplicate of the type stayed in training. One point per type (mean over its 10 runs), on the second test split; bars are group means, the dashed line is chance.}
\label{fig:cite_twins}\end{figure}

The same model, unchanged, on the CITE-seq data (RNA paired with surface protein, Section~\ref{sec:task}): 38 types on the same seven lineages after excluding the seven types with fewer than 200 cells, five deposited Liam seeds, mix plus every LOCTO type, both pair populations, 390 runs. Table~\ref{tab:cite} sets each cell-type-level claim of the main text beside its CITE-seq value. Figure~\ref{fig:cite_both_jobs} is the two-jobs reading on CITE-seq, the same picture as Figure~\ref{fig:type_level}A--B: the sampled arms sit inside the box, the deterministic network on its edge, and the aligned cosine clears the bar on both jobs in no run. Figure~\ref{fig:cite_twins} is the near-duplicate split described below. The task is harder (45 annotated types against 22) and the classifier is 0.006 to 0.014 AUROC behind the multiome. Everything else replicates in kind and close in size.

\begin{table}[htbp]
\centering
\caption{Every cell-type-level claim on both modality pairs. Multiome: 22 types, 230 runs; CITE-seq: 38 types, 390 runs. Novelty detection is on positives only, per run. Both-jobs shares are on cross-batch pairs. The learned-similarity rows are over ordered type pairs (462 multiome, 1{,}406 CITE-seq).}
\label{tab:cite}
\resizebox{\textwidth}{!}{\begin{tabular}{lcc}
\toprule
 & multiome & CITE-seq \\
\midrule
mix AUROC, cross-batch / within-batch pairs & 0.976 / 0.977 & 0.970 / 0.963 \\
error detection, $p_{\mathrm{sd}}$ / margin & 0.856 / 0.858 & 0.795 / 0.797 \\
gain at 50\% coverage, $p_{\mathrm{sd}}$ / margin & $+0.058$ / $+0.058$ & $+0.073$ / $+0.071$ \\
novelty detection, $p_{\mathrm{sd}}$ & 0.627 & 0.583 \\
\quad above 0.5 in & 19 of 22 types & 28 of 38 types \\
look-alike partner, median $p_{\mathrm{sd}}$ nearest / other & 0.057 / 0.0015 & 0.029 / 0.0006 \\
\quad AUROC nearest vs other, per run & 0.81 & 0.80 \\
learned vs embedding similarity, Spearman & $+0.68$ & $+0.54$ \\
\quad nearest neighbor identical & 16 of 22 & 24 of 38 \\
lineage mass at matched entropy, model / embedding & 0.573 / 0.568 & 0.676 / 0.672 \\
both jobs $>0.55$: MC dropout / ensemble / MFVI / deterministic & 88 / 82 / 69 / 53\% & 70 / 70 / 60 / 53\% \\
posterior over deterministic, novelty AUROC & $+0.055$ & $+0.034$ \\
\bottomrule
\end{tabular}}
\end{table}

Two things differ. The posterior's gain over the deterministic network on novelty detection is smaller on CITE-seq ($+0.034$ against $+0.055$). Within-batch and cross-batch pairs are ranked the same (AUROC 0.963 against 0.970) but thresholded differently at 0.5 (accuracy 0.941 against 0.922), which is calibration and not discrimination.

\paragraph{Near-duplicate types.}
The CITE-seq annotation contains pairs of types that differ by one marker, such as NK and NK CD158e1$^+$. Holding one of them out leaves its near-duplicate in training, and the model should not be surprised. Splitting the 38 held-out types by whether such a twin stayed in training (16 with, 22 without), the novelty AUROC of $p_{\mathrm{sd}}$ is 0.523 with a twin and 0.628 without on the test split, and 0.524 against 0.614 on a second test split built after the effect was seen (permutation $p = 0.002$ and $0.005$, 20{,}000 label permutations; Figure~\ref{fig:cite_twins}). The uncertainty rises on a type that is new and not on a type that is a relabeled neighbor. The multiome annotation has no such pairs, so this can only be seen here.


\clearpage
\section{Glossary of terms used in the manuscript and the accompanying code base}
\label{sec:glossary}

\paragraph{Assays and data.}
\begin{description}
\item[GEX] Gene expression, the count of transcripts per gene in a cell, from scRNA-seq.
\item[ATAC] Chromatin accessibility, the count of sequencing fragments per accessible region (peak) in a nucleus, from scATAC-seq. A region is accessible when it is depleted of nucleosomes.
\item[ADT] Antibody-derived tags: the abundance of surface proteins in a cell, measured by oligonucleotide-tagged antibodies in CITE-seq.
\item[Multiome] The 10x Genomics assay that measures GEX and ATAC in the same nucleus.
\item[CITE-seq] The assay that measures GEX and ADT in the same cell.
\item[Paired data] A dataset in which every cell is measured in two modalities and the pairing is known; the ground truth of the matching task.
\item[Modality] One kind of measurement of the cell (GEX, ATAC, ADT).
\item[Batch] One site--donor combination; the unit of technical variation. Thirteen in the multiome data, twelve in CITE-seq.
\item[Batch effect] The systematic difference between batches that is not biological.
\item[Cell type] The annotation of a cell in the benchmark; 22 types in the multiome data, 45 in CITE-seq.
\item[Lineage family] One of seven groups of cell types by hematopoietic lineage (Table~\ref{tab:families}), fixed before any family-level result.
\item[Relative] A cell type in the same lineage family as a given type.
\item[Liam] The variational autoencoder used to compute the frozen single-modality embeddings at the cell-type level and to initialize the encoders at the cell level; trained on one modality with no labels and no pairing.
\end{description}

\paragraph{Tasks and designs.}
\begin{description}
\item[Diagonal integration] Matching cells across modalities that were assayed in different cells.
\item[Cell-type level] The task of deciding whether two cells, one per modality, are of the same type.
\item[Cell level] The task of finding, for a cell in one modality, the same cell in the other.
\item[Match-modality task] The NeurIPS 2021 competition's cell-level task on the 20,009 held-out nuclei of site 4; its score is the average probability a method places on the true partner, which for a one-to-one assignment equals the accuracy.
\item[Pair] One RNA embedding and one ATAC (or ADT) embedding with the label ``same type'' or not.
\item[Within-batch / cross-batch pairs] Pairs whose two cells share a batch, or are built without regard to batch; two populations, never pooled.
\item[Mix] All cell types in training, validation and test.
\item[LOCTO] Leave-one-cell-type-out: one type removed from training and validation, all its cells in test.
\item[LOFO / LTFO] Leave-one-family-out / leave-two-families-out: one or two lineage families removed whole.
\item[Few-shot] LOFO with $k$ cells of the held-out family returned to training, on a test pool fixed across $k$.
\item[Seen / unseen] A cell type present in training, or held out from it.
\item[Challenge split] The competition's test set as the test set, everything else for training and validation.
\item[Challenge-LOCTO] The challenge split with one type removed from training and validation and the test set unchanged.
\item[Seed] One repetition of an experiment: its own data split and its own Liam embeddings or initialization. Five throughout.
\end{description}

\paragraph{Models and scores.}
\begin{description}
\item[Leg] The modality-specific part of the network, with its own weights.
\item[Shared block / projection] The part of the network both modalities pass through with the same weights; what makes the network siamese.
\item[Merged head] At the cell-type level, the layers that map the concatenation of the two representations to the pair probability.
\item[MFVI] Mean-field variational inference: a Gaussian posterior with independent components over the weights, trained by maximizing the evidence lower bound.
\item[Posterior placement] Which layers carry the posterior: the merged head only (\emph{base}) or every layer (\emph{legs}).
\item[Draw] One sample of the weights from the posterior; $S$ draws give $S$ predictions per input. $S = 200$ unless stated.
\item[DET / BNN / MCD / ENS] The deterministic network; the network with the MFVI posterior; the same network with Monte Carlo dropout at test time; a deep ensemble of five deterministic networks.
\item[Score matrix $S$] At the cell level, the scaled cosine similarity between every GEX cell and every ATAC cell of a test batch.
\item[Assignment $\pi$] The one-to-one matching of GEX to ATAC cells obtained by the Hungarian algorithm on the mean score matrix over seeds.
\item[Seed stability] The share of the five seeds whose own Hungarian assignment of cell $i$ disagrees with $\pi(i)$.

\item[$\bar p$, $p_{\mathrm{sd}}$] For one pair, the mean and the standard deviation of the $S = 200$ Monte-Carlo draws $p^{(s)}$ of the probability that the pair shares a type. $\bar p$ is the prediction; $p_{\mathrm{sd}}$ is the uncertainty score at the cell-type level.
\item[Confidence margin] $0.5 - |\bar p - 0.5|$, the distance of the prediction from the decision threshold. The deterministic control for $p_{\mathrm{sd}}$: any classifier has it without a posterior.
\item[Score matrix, temperature] At the cell level, $S_{ij} = \cos(z_i, z_j)/\tau$ between the $\ell_2$-normalized projections $z_i$ of a cell in one modality and $z_j$ of a cell in the other, with $\tau = 0.1$. The temperature sets how sharply the softmax in the loss concentrates on the best-matching candidate.
\item[InfoNCE] The symmetric contrastive loss over a block of $B$ paired cells, $\mathcal{L} = -\frac{1}{2B}\sum_{i=1}^{B}\Big[\log \frac{e^{S_{ii}}}{\sum_{j} e^{S_{ij}}} + \log \frac{e^{S_{ii}}}{\sum_{j} e^{S_{ji}}}\Big]$: each cell must rank its own partner above the other $B - 1$ cells in the block, in both directions.
\item[Hungarian assignment] The one-to-one assignment $\pi$ over a test batch that maximizes $\sum_i S_{i,\pi(i)}$ on the mean score matrix. Every cell in one modality is matched to exactly one cell in the other.
\item[Sinkhorn transport weight] The deterministic control at the cell level. Sinkhorn iterations turn $\exp(S/T)$ into a doubly stochastic matrix $P$; the score for cell $i$ is $P_{i,\pi(i)}$, the share of its unit mass the plan places on its assigned partner while every other cell competes for the same partners. $T$ is chosen as the best of nine values on the test set, which favors the control and is stated; $T = 1$ is reported beside it.
\item[$u_i$, top-1 vote] $u_i = \frac{1}{S}\sum_{s=1}^{S} \mathbf{1}\big[\arg\max_j S^{(s)}_{ij} \neq \pi(i)\big]$, the share of posterior draws ($S = 1{,}000$, 200 per seed) whose best match for cell $i$ is not its assigned partner.
\item[$v_i$, transport-plan score] $v_i = 1 - \frac{1}{S}\sum_{s=1}^{S} P^{(s)}_{i,\pi(i)}$, one minus the mean over draws of the weight each draw's own Sinkhorn plan places on the assigned pair, at its own temperature $T_v$.
\item[Row margin] The gap between the best and second-best entry of row $i$ of the mean score matrix; a naive control that ignores the one-to-one constraint.

\end{description}

\paragraph{Measures.}
\begin{description}

\item[ECE] Expected calibration error over $B = 15$ equal-width bins of $\bar p$, $\mathrm{ECE} = \sum_{b=1}^{B} \frac{n_b}{N}\,\big|\mathrm{acc}_b - \mathrm{conf}_b\big|$, where $n_b$ is the number of pairs in bin $b$, $\mathrm{acc}_b$ their accuracy, and $\mathrm{conf}_b$ their mean $\bar p$. Zero means the predicted probability matches the observed frequency in every bin.
\item[Lift over a blind rule] At a fixed coverage $c$, the share of errors (or of the held-out type's pairs) among the discarded predictions, divided by $1 - c$, which is what discarding at random removes. A lift of 1.0 means the score did no better than chance.
\item[Entropy-matched lineage mass] For a held-out type $T$, the row $S[T, \cdot]$ of the learned similarity matrix, normalized to a distribution over seen types, and the share of it on $T$'s own lineage family. The embedding's row is raised to the power that gives it the model row's entropy before the two shares are compared, so that peakiness alone cannot produce a difference.
\item[Spearman correlation] Rank correlation between two scores over pairs or over type pairs; 1 means the same ordering.

\item[Brier score] Mean squared error of the predicted probability against the label over $N$ pairs, $\mathrm{Brier} = \frac{1}{N}\sum_{i=1}^{N} (\bar p_i - y_i)^2$, with $y_i \in \{0, 1\}$. A proper scoring rule: it is minimized only by the true probability, so a lower value means a better probability and not a better threshold. Range 0 to 1.

\item[NLL] Negative log-likelihood of the labels under the predicted probabilities, $\mathrm{NLL} = -\frac{1}{N}\sum_{i=1}^{N} \big[\, y_i \log \bar p_i + (1 - y_i) \log (1 - \bar p_i) \,\big]$. Also a proper scoring rule; unlike the Brier score it is unbounded and punishes a confident wrong prediction without limit, so it is the more sensitive of the two to overconfidence.

\item[Matching accuracy] The share of cells whose assigned partner is their own.
\item[Error detection] The AUROC of an uncertainty score for separating wrong predictions from correct ones.
\item[Same-type / relative / other-family error] A wrong match whose partner is of the query's own type, of a type in the same lineage family, or of a type in another family.
\item[Abstention] Discarding the least certain share of predictions by a score; reported as the accuracy on what is kept and the share of errors removed, against discarding at random.
\item[Coverage] The share of predictions kept after abstention.
\item[Novelty detection] Within a held-out run, the AUROC of a score for separating the held-out type's cells from seen cells.
\item[Permuted control] The same share recomputed after the assigned partners are shuffled within the batch; the level a lineage share must exceed to mean anything.
\item[Pre-registration] A prediction and a bar written down before a run, and reported whether or not the bar was met.
\end{description}

\end{document}